\documentclass{article}

\usepackage{PRIMEarxiv}

\usepackage[utf8]{inputenc} 
\usepackage[T1]{fontenc}    
\usepackage{hyperref}       
\usepackage{url}            
\usepackage{booktabs}       
\usepackage{amsfonts}       
\usepackage{amssymb}    
\usepackage{amsthm}     
\usepackage{mathtools}  
\usepackage{tikz}       
\usepackage{caption}    
\usepackage{natbib}     
\usepackage{subcaption} 
\usepackage{multicol}   
\usepackage{nicefrac}       
\usepackage{microtype}      
\usepackage{lipsum}
\usepackage{fancyhdr}       
\usepackage{graphicx}       
\graphicspath{{media/}}     

\newtheorem{definition}{Definition}
\newtheorem{example}{Example}
\newtheorem{theorem}{Theorem}
\newtheorem{proposition}{Proposition}
\newcommand{\comment}[1]{}

\title{Action Languages Based Actual Causality for Computational Ethics: a Sound and Complete Implementation in ASP
}

\author{
  Camilo Sarmiento \\
  Sorbonne Université, CNRS, LIP6 \\
  F-75005 Paris, France \\
  \texttt{camilo.sarmiento@lip6.fr} \\
  \And
  Gauvain Bourgne \\
  Sorbonne Université, CNRS, LIP6 \\
  F-75005 Paris, France \\
  \texttt{gauvain.bourgne@lip6.fr}\\
  \And
  Katsumi Inoue \\
  National Institute of Informatics \\
  Tokyo, Japan \\
  \texttt{inoue@nii.ac.jp}\\
  \And
  Daniele Cavalli \\
  École Normale Supérieure -\\
  Université PSL,\\
  République des savoirs,\\
  Collège de France, CNRS\\
  F-75005 Paris, France\\
  \texttt{daniele.cavalli@ens.psl.eu}\\
  \And
  Jean-Gabriel Ganascia \\
  Sorbonne Université, CNRS, LIP6 \\
  F-75005 Paris, France \\
  \texttt{jean-gabriel.ganascia@lip6.fr}\\
}

\begin{document}

    \maketitle

    \begin{abstract}
    Although moral responsibility is not circumscribed by causality, they are both closely intermixed. Furthermore, rationally understanding the evolution of the physical world is inherently linked with the idea of causality. Thus, the decision-making applications based on automated planning inevitably have to deal with causality---especially if they consider imputability aspects or integrate references to ethical norms. The many debates around causation in the last decades have shown how complex this notion is and thus, how difficult is its integration with planning. As a result, much of the work in computational ethics relegates causality to the background, despite the considerations stated above. This paper's contribution is to provide a complete and sound translation into logic programming from an actual causation definition suitable for action languages---this definition is a formalisation of Wright’s NESS test. The obtained logic program allows to deal with complex causal relations. In addition to enabling agents to reason about causality, this contribution specifically enables the computational ethics domain to handle situations that were previously out of reach. In a context where ethical considerations in decision-making are increasingly important, advances in computational ethics can greatly benefit the entire AI community.
\end{abstract}
    
    \section{Introduction}
    
\begin{figure}[htb]
    	\centering
    	\resizebox{14.5cm}{!}{\input{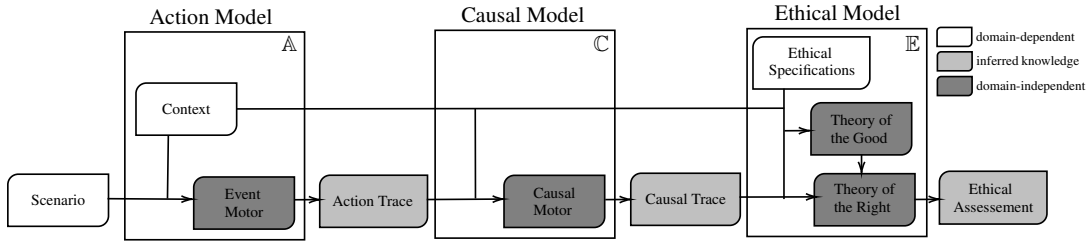}}
    	\caption{Modular framework for representing and applying ethical principles.}
    	\label{image:ACE_structure} 
    \end{figure}

    We aim to design an ethical supervisor that can be embedded in agents so that their actions obey several moral prescriptions. The risks associated with their actions are then limited. The need for such a supervisor arises from the growing number of agents capable of handling tasks with increasing complexity to whom growing responsibilities are delegated \citep{tolmeijer_implementations_2021}. As a result, the space they occupy in our daily lives is growing, as is the risk associated with this trend. The agents we are interested in are technical objects capable of taking into account the information that they receive from their environment and act to modify it. This definition should not be confused with the one used in the philosophy of action \citep{schlosser_agency_2019} and human science in general, where the concept of agent is usually linked to agency. The ethical supervisor that we use is the $\mathbb{ACE}$ ($\mathbb{A}$ction-$\mathbb{C}$ausality-$\mathbb{E}$thics) modular framework presented in Figure \ref{image:ACE_structure}. It is the logical continuation of a series of works \citep{berreby_modelling_2015,berreby_declarative_2017,berreby_event-based_2018,bourgne_ace_2021}.
    
    The purpose of this paper is first to explain that when designing such a supervisor, it is essential to explore imputability---`no blame or praise may be assigned without some account of causal relationship between an agent and an outcome' \citep{beebee_oxford_2009}. Logically, the second purpose of this paper is to provide a mechanism allowing the exploration of causality suitable for decision-making. This inevitably leads us to look at causation as the major consideration of this paper. Because of its essential role in human reasoning---both in trivial and in complex situations---numerous works in a variety of disciplines have unsuccessfully tried to propose a widely agreed upon theory of causation. Since we are in an operational framework given that our focus is on ethical decision-making, we can make a couple of assumptions while remaining relevant. Therefore, we place ourselves in a classical planning framework, which assumes problems are discrete and deterministic. Unlike \textit{type causality}---which seeks to determine general causal relationships---\textit{actual causality} fits our purpose because it is concerned with particular events \citep{halpern_actual_2016}. Limiting ourselves to a simplified framework and to actual causality does not make causality trivial, many issues still remain.
    
    This paper is structured as follows. Section \ref{sec:Probleme} shows how important the ability to establish causal relationships is when modelling ethical reasoning. Despite encouraging initial results in the field of computational ethics, much remains to be done. We argue that this gap is partly due to limitations that can be explained by the lack of work incorporating mechanisms for establishing complex causal relationships. Section \ref{sec:Modele} introduces the action language semantics---in which we encode causal knowledge---allowing the concurrency of events. Section \ref{sec:Causal} discusses what is the appropriate approach to causation for our causal inquiry. In this section we explore two main highly influential theories of causation: \textit{regularity} and \textit{counterfactual}. Section \ref{sec:Actual_cause} offers a more comprehensive description of the actual causality definition proposed in \citep{sarmiento_action_2022-1}. This definition is factual and independent of policy choices and allows to handle complex cases of causality in ethical decision-making applications. Section \ref{sec:Logic_program} presents the translation from our actual causality definition into logic programming, while stating the soundness and completeness Theorems. Finally, we conclude and give some perspectives in Section \ref{sec:conclude}. Proofs of the Theorem and Propositions are given in Appendix. This paper is an extended version of \citep{sarmiento_action_2022-1} in which the significant new elements are: the discussion in Section \ref{sec:Probleme}, the example on pollution used throughout the article, Propositions \ref{pro:direct_NESS_literal} to \ref{pro:non_unicity} which characterise our causal relations definitions, a more comprehensive description of the causality definitions, and a translation from our actual causality definition into logic programming which we prove to be sound and complete in Theorems \ref{the:complete_and_sound} to \ref{the:complete_and_sound_2}.

    \section{Causality in Computational Ethics}
\label{sec:Probleme}
    
    In recent years, various works in the field of computational ethics have shown that it is possible to formalise moral prescriptions---and this by means of a large number of techniques \citep{tolmeijer_implementations_2021}. Primarily, those belonging to the consequentialist and deontological traditions, which allow a better formalisation than more recent theoretical reflections (e.g ethics of care, quantum ethics). To do this, these works have mainly relied on thought experiments, such as the trolley problem introduced by \cite{foot_problem_1967}. The first challenge was to formalise the problem---i.e. representing it in a computer language so as to reproduce the event sequence of the thought experiment. For the trolley problem, it was necessary to represent the fact that inaction would cause the death of five people, while the action of diverting the trolley would only cause the death of a single person.
    
    The second challenge was to formalise moral prescriptions used for assessing the acceptability of possible action choices in the thought experiment. Various moral prescriptions have been formalised so far. Following are some of the prescriptions belonging to consequentialism with the works where they have been formalised: \textit{prohibiting purely detrimental actions}, an action is impermissible if its consequences are purely bad \citep{berreby_declarative_2017}; \textit{principle of benefits vs. costs}, an action is impermissible if its bad consequences are more significant than its good consequences \citep{berreby_declarative_2017,bourgne_ace_2021}; \textit{principle of least bad consequence}, an action is impermissible if its worst consequence is considered the least good of all possible alternative consequences \citep{ganascia_non-monotonic_2015,berreby_declarative_2017,lindner_hera_2017}; \textit{Pareto principle}, an action is impermissible if there is another possible action whose positive consequences are better or whose negative consequences are less bad \citep{lindner_hera_2017}; \textit{act utilitarianism}, an action is impermissible if there is another action with consequences that are better overall \citep{berreby_declarative_2017,lindner_hera_2017,bourgne_ace_2021}; \textit{rule utilitarianism}, an action is impermissible if there is another action for which the universalisation of the rule to which it conforms produces more utility than the universalisation of the first action corresponding rule \citep{berreby_declarative_2017}. Following are some of the moral prescriptions belonging to deontology with the works where they have been formalised: \textit{codes of conduct}, an action is impermissible if it goes against a prohibition in the code followed \citep{berreby_declarative_2017,limarga_non-monotonic_2020}; \textit{\citeauthor{kant_fondements_2013}'s formula of the end in itself} \citep{kant_fondements_2013}, an action is impermissible if it has consequences---which involve an individual---other than the ends of the action \citep{berreby_declarative_2017,lindner_formalization_2018,bourgne_ace_2021}; \textit{\citeauthor{daquin_summa_1266}'s doctrine of double effect} \citep{daquin_summa_1266}, an action is impermissible if at least one of these four conditions is met: (i) the action is intrinsically bad, (ii) at least one bad consequence is a means to a good one, (iii) at least one bad consequence was intended, (iv) its bad consequences are greater than its good consequences \citep{berreby_modelling_2015,berreby_declarative_2017,govindarajulu_automating_2017,lindner_hera_2017,bonnemains_embedded_2018}. The diversity represented by these few examples is proof that the formalisation of moral prescriptions is possible and not limited to a small family of ethical systems.
    
    Despite these encouraging results, there is still much room for improvement. Indeed, what makes ethical reasoning so laborious to reproduce is that ethics lie in the subtlety of each problem. Thus, the complexity of the task facing computational ethics is as much in the formalisation of moral prescriptions as it is in the formalisation of the problem---specifically in the integration of all the subtleties relevant to the problem into its formalisation. Consider the situation where an agent has to decide whether the action of making a vaccine compulsory is permissible. From that perspective, the conflict to be resolved will be primarily between the right to personal liberty and the right to collective safety \citep{united_nations_general_assembly_universal_1948}. Undoubtedly, the ethical reasoning will change if a significant part of the population is allergic to the vaccine. Moreover, the ethical considerations will change radically if the group of allergic people is mainly composed of individuals of the same gender or ethnicity. If any agent is to make this decision based on an ethical decision-making mechanism, it must be able to take into account the subtleties relevant to each of these cases. The solutions currently proposed are still far from being able to handle the complexity involved in real world situations.
        
    The main argument in this section is that the current limitations of existing solutions are mainly due to the systematic absence---with a few exceptions \citep{berreby_modelling_2015,berreby_declarative_2017,bourgne_ace_2021,lindner_hera_2017,lindner_formalization_2018}---of a mechanism for establishing causal relationships. The importance of causality appears obvious when the moral prescriptions belong to consequentialism, i.e. the theory that the acceptability of actions depends exclusively on the value of its consequences. Paradoxically, we shall see that causality is rarely given the place it deserves. Although the link seems less intuitive, causality is also essential in some theories of deontological ethics. This is the case when considering whether an action is a means to an end in \citeauthor{kant_fondements_2013}'s second formulation of the categorical imperative \citep{kant_fondements_2013} or \citeauthor{daquin_summa_1266}'s doctrine of the double effect \citep{daquin_summa_1266}. Because of its importance in most of the formalised moral prescriptions, causality's absence has two main consequences on the proposals made so far: (i) the oversimplification of the problem formalisation and (ii) the fact of leaving out problems---as Examples \ref{ex:preemption}, \ref{ex:duplication}, and \ref{ex:switches}---that may contain overdetermination \citep{wright_causation_1985}:
        
        \begin{quote}
            \textit{overdetermined causation}: cases in which a factor other than the specified act would have been sufficient to produce the injury in the absence of the specified act, but its effects either (1) were preempted by the more immediately operative effects of the specified act or (2) combined with or duplicated those of the specified act to jointly produce the injury.
        \end{quote}
    
    (i) The oversimplification of the problem arises from the lack of ability to create complex causal links. Hence, the only way to link an action to a consequence is either to formalise it as an intrinsic effect of the action, or to assess an initial and an end state and attribute all changes to the action. Considering the trolley problem, the first approach corresponds to formalising the action `pulling the switch lever' as having the intrinsic effect of killing the individual on the second track while the action `doing nothing' results in the death of the five individuals on the main track \citep{limarga_non-monotonic_2020,bonnemains_embedded_2018}. The second approach takes the initial state where all individuals are alive, compares it to the final state where some individuals are dead, and considers that all changes are effects of one of the two possible actions performed in between. These shortcuts have appeared to be viable solutions because the problems studied are mostly thought experiments---designed to help us understand the asymmetry of our judgement in certain situations, such as why it would be acceptable to sacrifice one person to save five in one context but not in another that is very similar \citep{nyholm_ethics_2016}. Accordingly, in the studied problems the number of factors that can be taken into consideration is very small, the outcomes are certain, and the number of possible actions is very limited, a context far from that of the real world. This apparent simplicity has therefore masked the complexity of one of the challenges of the field: capturing the subtlety of each problem in its formalisation. These shortcuts are particularly inappropriate as they may have a negative influence on the ethical assessment of the problem. Given the first approach formalisation of the two possible actions in the trolley problem, these are considered intrinsically bad in relation to the right to live because one of their intrinsic effects is the death of one or more individuals. The act of `pulling the switch lever' would be in some way equated to shooting someone. The reader will quickly understand that this is not the case and that it is a product of the formalisation. The second approach does not have this problem. However, we can question its capacity to properly manage imputability. Consider a case where one of the individuals in the trolley problem is shot by a hitman before the train diverts. The attribution of this death to the agent being able to divert the trolley is problematic. Yet, this is what happens if we reduce causality to a simple comparison of states. These imputation errors become unavoidable when other agents can act on the world and we consider possible to have concurrent actions---situations that are far from being exceptional and that we cannot simply ignore. The solution to both approaches undesirable effects is to formalise the dynamics of the problem. For instance, it would be more appropriate to formalise the two actions of the trolley problem as having an intrinsic effect on the movement of the trolley and not on the lives of the individuals. By contrast with commonly used approaches, in this context, the agent's actions are no longer directly linked to the death of the individuals which leads to the essential need of a mechanism establishing causal relationships.
    
    (ii) The reasons for not dealing with overdetermination are the arising difficulties that can only be handled by examining causality. Cases of overdetermination are commonly found in the complexity of reality (cause of pollution, cause of suicide, cause of economic loss, \dots), they give rise to numerous questions in law---for example the Paradox of \textit{Conditio Sine Qua Non} discussed in \citep{satoh_disjunction_2006}. They have therefore been the subject of much work in the fields of causation, whether by lawyers, philosophers or mathematicians. Insofar as overdetermination is a phenomenon that an agent in a real context may face, the inability of all the propositions mentioned so far to deal with these cases is an important limitation of the field. In addition, to avoid these cases, propositions leave aside concurrency of events, which eliminates many common cases an agent may face. Thus, to achieve the desired goal it is not only necessary that the proposals in the field incorporate a mechanism for establishing causal relationships, but also that this mechanism is sufficiently complex to handle cases of causal overdetermination.
    
    The examples below will be used throughout this paper. We may seek to explore these examples as they present different cases of causal overdetermination while illustrating ethical issues.
   
    \begin{example}[\textit{pollution---preemptive causation}]\label{ex:preemption}
        A village along a river is home to $n$ families. The drinking water used by the inhabitants of the village comes from a water treatment plant that draws water from the river, which in turn comes from a lake located upstream of the river. However, the capacity of this plant is limited, it can only treat water if it has a pollution indicator below a threshold. When water from the mountain reaches the lake, the pollution indicator is zero. There are two potential sources of pollution to the lake: (i) industrial wastewater from a factory that produces connected speakers for a famous online shopping site ($w_s$) and (ii) industrial wastewater from a factory that produces life-saving medicines for $k$ patients ($w_m$). Under normal circumstances, $w_m$ does not pollute the lake because it is treated by a wastewater treatment plant before being discharged into the lake. We assume that the discharges are necessary for the launching operation of the factory and that the production management from both factories is ensured by automated agents. We consider the scenario where the agent in charge of the connected speakers factory launches the production. The discharged wastewater increases the pollution indicator to the threshold. It turns out that the plant treating the wastewater from the medicine factory has been out of order since the beginning of the month. A few hours after the launching of the connected speakers factory, the agent in charge of the medicines factory launches the production. The discharged wastewater increases the pollution indicator to two times the threshold. The inhabitants of the village are left without water in this scenario.
    \end{example}
    
    Example \ref{ex:preemption} is a case of preemptive causation. The production of speakers and the production of medicines---given the broken state of $w_m$ treatment plant---are both sufficient to cause the harm. Indeed, individually each discharge would raise the level to the threshold at which the village inhabitants would be left without water. However, the eventuality that $w_m$ discharges deprive the inhabitants of water is preempted by the anteriority of $w_s$ discharges. In other words, effects of $w_s$ took precedence over those of $w_m$. Therefore, the production of speakers is identified as a cause of the harm.
    
    \begin{example}[\textit{pollution---duplicative causation}]\label{ex:duplication}
        We remain in the same framework as the Example \ref{ex:preemption}, except that in this scenario the two agents launch the production at the same time. The discharged wastewater increases the pollution indicator to two times the threshold. The inhabitants of the village along the river are left without water in this scenario.
    \end{example}
    
    Example \ref{ex:duplication} is a case of duplicative causation. We remain in the same situation as above, except that this time the effects of $w_s$ and $w_m$ discharges on water quality occur simultaneously---they combine to produce the harm jointly. The production of speakers and medicines are both identified as causes of the harm.
    
    Besides illustrating two types of overdetermination, these examples may well show the complexity behind ethical assessment. Indeed, many factors make the problem more complex. First, we have the fact that the value produced by the polluting activity is different. On one hand, the production of a so-called comfort good, on the other hand, the production of a medicine essential for the survival of individuals. We could add that the speakers factory supports the region by giving work to a large part of the inhabitants of the village, whereas the medicine factory has completely automated its production line. We could also add that the discharges from one factory are authorised, while those from the other are not. In the end, all these factors can have a significant influence on the permissibility of each action. However, all this richness can only be explored if one has sufficient expressiveness to formalise it and a mechanism to establish sufficiently complex causal relationships. We will start by introducing the action language with which we address the first.

    \section{Action Language Semantics}
\label{sec:Modele}
    
    The purpose of our action language is to determine the evolution of the world given a set of actions corresponding to deliberate choices of the agent. Those actions might trigger some chain reaction through external events. As a result, we need to keep track of both: the state of the world and the occurrence of events---the term `event' connoting `the possibility of agentless actions' \citep[chap 12]{russell_artificial_2010}. This task is the simplest kind of temporal reasoning---temporal projection. Different action languages allowing temporal projection have been proposed such as PDDL \citep{ghallab_pddl_1998,haslum_introduction_2019} and action description languages $\mathcal{A}$, $\mathcal{B}$, and $\mathcal{C}$ \citep{gelfond_action_1998}. However, the semantics of $\mathcal{A}$ \citep{gelfond_representing_1993}, $\mathcal{B}$, and PDDL deterministic fragment---corresponding to ADL \citep{pednault_adl_1989}---do not allow concurrency of events. To have a semantics that takes into account concurrency it is necessary to jump directly to $\mathcal{C}$ \citep{giunchiglia_action_1998} or PDDL+ \citep{fox_modelling_2006} which semantics is adapted respectively to non deterministic actions or durative actions, thus inconsistent with either our deterministic actions assumption or our discrete time assumption. The advanced state of maturity of PDDL \citep{ghallab_pddl_1998,haslum_introduction_2019}, its vocation to facilitate interchangeability, and its use by a large community, are all meaningful arguments in favour of this formalism---gradually extended by different fragments. We therefore base our approach on an action language whose semantics is an intermediate point between the deterministic fragment of PDDL and PDDL+. This formalism works on a decomposition of the world into two sets: $\mathbb{F}$ corresponding to variables describing the state of the world, more precisely ground fluents representing time-varying properties; $\mathbb{E}$ representing variables describing transitions, more precisely ground events that modify fluents.
    
    A fluent literal is either a fluent $f\in \mathbb{F}$, or its negation $\neg f$. We denote by $Lit_{\mathbb{F}}$ the set of fluent literals in $\mathbb{F}$, where $Lit_{\mathbb{F}} = \mathbb{F}\cup\left\{\neg f|f\in \mathbb{F} \right\}$. The complement of a fluent literal $l$ is defined as $\overline{l}=\neg f$ if $l=f$ or $\overline{l}=f$ if $l=\neg f$. By extension, for a set $L\subseteq Lit_{\mathbb{F}}$, we have $\overline{L} = \left\{\overline{l},l\in L\right\}$. 
        
        \begin{definition}[state $S$]\label{def:state}
            The set $L\subseteq Lit_{\mathbb{F}}$ is a state iff it is:
                \begin{itemize}
                    \item Coherent: $\forall l\in L, \overline{l}\not\in L$.
                    \item Complete: $\left|L\right| = \left|\mathbb{F}\right|$, i.e. $\forall f\in \mathbb{F}, f\in L$ or $\neg f \in L$.
                \end{itemize}
        \end{definition}
    
    A complete and coherent set of fluent literals thus determines the value of each of the fluents. An incoherent set cannot describe a reality. However, in the absence of information or for the sake of simplification, we can describe a problem through a coherent but incomplete set. We will call such a set a partial state. We model time linearly and in a discretised way to associate a state $S_t$ to each time point $t$ of a set $\mathbb{T} = \left\{-1,0,\dots,N\right\}$. Having a bounded past formalisation of a real problem, we gather all states before $t=0$---time point to which corresponds the state $S_0$ that we call initial state---in an empty state $S_{-1} = \varnothing$.
    
    We place ourselves in a framework of concurrency where $E_t$ is the set of all events which occur at a time point $t$. Therefore, $E_t$ is what generates the transition between the states $S_t$ and $S_{t+1}$. Thus, the states follow one another as events occur, simulating the evolution of the world. $E_{-1}$ is the set that gathers all events which took place before $t=0$, such that $E_{-1}=\left\{ini_l,l\in S_0\right\}$. Events are characterised by two elements: preconditions give the conditions that must be satisfied by the state in order for them to take place; effects indicate the changes to the fluents that are expected to happen if they occur. The preconditions and effects are respectively represented as formulas of the language $\mathcal{P}$ and $\mathcal{E}$ defined as follows:
    $$\mathcal{P} \Coloneqq l|\psi_1 \wedge \psi_2|\psi_1 \vee \psi_2 \hspace{0.8cm} \mathcal{E} \Coloneqq l|\varphi_1 \wedge \varphi_2$$
    where $l\in Lit_{\mathbb{F}}$ and the logical connectives $\wedge$, $\vee$ have standard first-order semantics. We can then deduce that if $\varphi\in\mathcal{E}, ~\varphi=\bigwedge_{i\in \left\{1,\dots,m\right\}} l_i$. For the sake of brevity, we adopt a set notation for $\varphi \in \mathcal{E}$ which we will use where relevant, such that $\varphi = \left\{l_i,~i\in \left\{1,\dots,m\right\}\right\}$. Note that conditional effects were not kept for implementations reasons. The \citeauthor{nebel_compilability_2000}'s compilation schemes \citep{nebel_compilability_2000} allow us to deal with the same problems. We denote $pre$ and $eff$ the functions which respectively associate preconditions and effects with each event: $pre: \mathbb{E} \mapsto \mathcal{P}$, $eff: \mathbb{E} \mapsto \mathcal{E}$. Given the expression of $E_{-1}$, the application of $eff$ to each element of the set is $eff(ini_l)=l$ with $l\in S_0$, thus $eff(E_{-1})=S_0$. Moreover, given a formula $\psi\in\mathcal{P}$ and a partial state $L$, $L\vDash\psi$ is defined classically: $L\vDash l$ if $l \in L$, $L \vDash \psi_1 \wedge \psi_2$ if $L \vDash \psi_1$ and $L \vDash \psi_2$, and $L \vDash \psi_1 \vee \psi_2$ if $L \vDash \psi_1$ or $L \vDash \psi_2$.
    
    Our work is a logical continuation of works such as \citep{hopkins_causality_2007,berreby_event-based_2018,batusov_situation_2018,leblanc_explaining_2019}, who attempted to link action languages and causation. To the best of our knowledge, \cite{batusov_situation_2018} and \cite{berreby_event-based_2018} are the first to give a definition of actual cause in action languages. However, each work has its own limitations that we try to address. In \citeauthor{batusov_situation_2018}'s paper \citep{batusov_situation_2018}, many working perspectives are mentioned:
    
    \begin{quote}
        It is clear that a broader definition of actual cause requires more expressive action theories that can model not only sequences of actions, but can also include explicit time and concurrent actions. Only after that one can try to analyze some of the popular examples of actual causation formulated in philosophical literature. Some of those examples sound deceptively simple, but faithful modelling of them requires time, concurrency and natural actions.
    \end{quote}

    At the moment, the proposed action language tackles both concurrency and time---at least discrete time. We will now introduce `natural actions' that we denote exogenous events. These events are what distinguish our proposal from $\mathcal{A}_c$ \citep{baral_reasoning_1997}---the allowing concurrency version of $\mathcal{A}$. The set $\mathbb{E}$ is divided into two sub-sets: $\mathbb{A}$, which contains the actions carried out by an agent and thus subjected to a volition; $\mathbb{U}$, which contains the exogenous events---equivalent to \texttt{:event} in PDDL+ \citep{fox_modelling_2006} and triggered axioms in Event Calculus \citep{mueller_commonsense_2014}---which are triggered as soon as all the $pre$ are fulfilled, therefore without the need for an agent to perform them. Thus, for exogenous events triggering conditions and preconditions are the same. In contrast, the triggering conditions for actions necessarily include preconditions but those are not sufficient. The triggering conditions of an action also include the volition of the agent or some kind of manipulation by another agent. To keep track of these subtleties that could be relevant in the causal inquiry we introduce triggering conditions represented as formulas of the language $\mathcal{P}$. We denote $tri$ the function which associates triggering conditions with each event: $tri: \mathbb{E} \mapsto \mathcal{P}$.
    
    The occurrence of events $(e,t)\in\mathbb{E}\times\mathbb{T}$ and $(e',t)\in\mathbb{E}\times\mathbb{T}$ in the state $S_t$ is said to be interfering if the set $\left\{l, l\in eff(e)\cup eff(e')\right\}$ is not coherent according to Definition \ref{def:state}.
    
    \begin{definition}[context $\kappa$]
        Given an initial state $S_0$, the context denoted as $\kappa$ is the octuple $(\mathbb{E},\mathbb{F},pre, tri,eff,S_0,>,\mathbb{T})$, where $>$ is a partial order which represents priorities that ensure the primacy of one event over another when both are interfering.
    \end{definition}
    
    As mentioned earlier, effects indicate the changes to the fluents that are expected to happen if an event occurs. Because of the complexity of reality, it may turn out that causally an event has more or less effects than those attributed by $\mathcal{E}$. Let us take the example of an agent who wants to turn on a light by pressing a switch. In a first scenario, it is possible that the agent's action causes an overheating in the electrical circuit and triggers a fire. When formalising the action of switching on the light, it is not intuitive to take into account the overheating and then the fire as intrinsic effects. Besides affecting the generality of the formalisation, we previously mentioned that this could influence the ethical evaluation. In these cases, we will prefer to break down the process by introducing exogenous events. In the above fire example, we will therefore have an exogenous event corresponding to a fire outbreak---an agentless event---which will be triggered when a defective circuit is present and the switch is pressed. We are therefore in the presence of a causal chain. The cases where the action has more effects than those with which it has been formalised are typical cases where causality is necessary. In a second scenario, it may happen that the agent performs the action but that the expected effects are not produced simply because the light was already on. This does not prevent the action from being performed, and we want to keep a trace of the event without having to consider that its effect has taken place. This is especially the case if the action has several effects and only one of them does not actually occur. This second scenario is possible because of the common sense law of inertia, i.e. we consider that all fluent literals are inertial, which means we assume their truth value remains the same from a state to another unless the occurrence of an event changes that. On the opposed, if fluent literals were non inertial, a fluent literal true at $t$ will be false at $t+1$ unless its maintenance is specified by an occurrence of event. This second case can be resumed as cases where some of the fluents of the state have already the value attributed by an effect. Since the effects that an event had at the time it occurred is a basic causal information on which we will rely---inextricably linked to imputability---it is important to keep track of them.
    
    \begin{definition}[actual effects $actualEff(E,L)$]\label{def:actualEff}
        Given a context $\kappa$, the predicate $actualEff(E,L)$ which associates a set of events $E\in \mathbb{E}$ given a partial state $L$, to a partial state representing the actual effects of $E$ if occurring in the state $L$, is defined as:
        
        \begin{align*}
            actualEff(E,L) &= \bigcup_{e\in E} actualEff(\left\{e\right\},L)\\
            &= \left\{l_i,\exists e\in E,l_i\in eff(e),\text{and}~l_i\notin L\right\}
        \end{align*}
    \end{definition}
    
    For the sake of conciseness we adopt an update operator giving the resulting state when an event occurs at a given state.
    
    \begin{definition}[update operator $\triangleright$]\label{def:update_op}
        Given a context $\kappa$ and set of events $E\in \mathbb{E}$, the update operator which we use as follows $S_t\triangleright E$ expresses $S_t\setminus \overline{actualEff(E,S_t)}\cup actualEff(E,S_t)$.
    \end{definition}
    
    The information given by $actualEff(E,L)$ and $\triangleright$ can be equated to basic causal information given by the evolution of the world. Besides being causal, this information is directional since it is inconceivable in our semantics to say that the actual effect of the event is the cause of it. Therefore, we can rely on the events that occur and their actual effects to simulate the evolution of the world from the initial state $S_0$.
    
    \begin{definition}[induced state sequence $\mathcal{S}_{\kappa}$]
        Given a context $\kappa$ and a sequence of events $\epsilon = E_{-1},E_0,\dots,E_n$, such that $n\leq |\mathbb{T}|$, the induced state sequence of $\epsilon$ is a sequence of complete states: $\mathcal{S}_{\kappa}(\epsilon) = S_0,S_1,\dots,S_{n+1}$ such that $\forall t \in \left\{-1,\dots,n\right\}, ~S_{t+1} = S_t\triangleright E_t$.
    \end{definition}
    
    Though this can be defined for every $\epsilon$, not all $\epsilon$ are possible given (i) the need to satisfy preconditions, (ii) the concurrency of events that must respect priorities, and (iii) the triggering of events that must respect priorities too.
    
    \begin{definition}\label{def:valid}
        Let $\epsilon$ be a sequence of events $\epsilon = E_{-1},E_0,\dots,E_n$, such that $n\leq |\mathbb{T}|$, and let's denote by $\mathcal{S}_{\kappa}(\epsilon) = S_0,S_1,\dots,S_{n+1}$ its induced state sequence. We shall say that $\epsilon$ is:
        \begin{itemize}
            \item Executable in $\kappa$: if $~\forall t \in \left\{0,\dots,n\right\}, ~S_t \vDash pre(E_t)$.
            \item Concurrent correct with respect to $\kappa$: $\neg\exists (e,e')\in E_t^2, ~e>e'$.
            \item Trigger correct with respect to $\kappa$: if $~\forall t \in \left\{0,\dots,n\right\}, ~\forall e'\in \mathbb{E}$ such that $~S_t \vDash tri(e')$, then $e' \in E_t$ or $\exists e\in E_t, ~e>e'$.
            \item Valid in $\kappa$: if and only if, executable in $\kappa$, concurrent correct with respect to $\kappa$, and trigger correct with respect to $\kappa$.
        \end{itemize}
    \end{definition}
    
    Finally, if we consider only a set of timed actions as an input which we call scenario, we have:
    
    \begin{definition}[traces $\tau_{\sigma,\kappa}^e$ and $\tau_{\sigma,\kappa}^s$]\label{def:traces}
        Given a scenario $\sigma \subseteq \mathbb{A}\times \mathbb{T}$ and a context $\kappa$, the event trace $\tau_{\sigma,\kappa}^e$ of $\sigma,\kappa$ is the sequence of events $\epsilon = E_{-1},E_0,\dots,E_n$ valid in $\kappa$, such that: $\forall t \in \left\{0,\dots,n\right\}, \forall e\in E_t, ~e \in \mathbb{A} \Leftrightarrow (e,t) \in \sigma$. Its induced state sequence is the state trace $\tau_{\sigma,\kappa}^s$.
    \end{definition}
    
    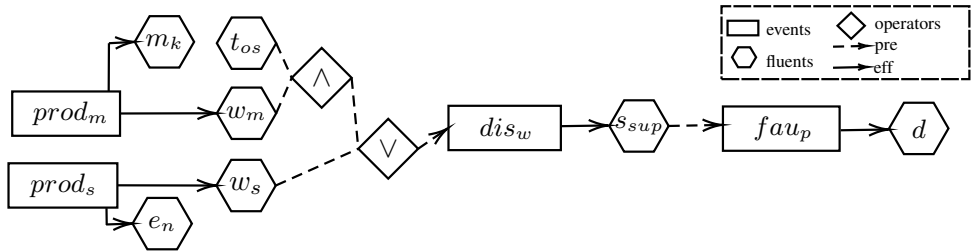
\begin{figure}[b]
        \centering
    	\resizebox{13cm}{!}{\tikzset{every picture/.style={line width=0.75pt}} 

\begin{tikzpicture}[x=0.75pt,y=0.75pt,yscale=-1,xscale=1]

\draw   (65,65.16) -- (113.6,65.16) -- (113.6,85.16) -- (65,85.16) -- cycle ;
\draw   (261.67,71.63) -- (312.33,71.63) -- (312.33,91.63) -- (261.67,91.63) -- cycle ;
\draw   (385.4,71.93) -- (438,71.93) -- (438,91.93) -- (385.4,91.93) -- cycle ;
\draw   (146,124.78) -- (139.31,136.37) -- (125.93,136.37) -- (119.24,124.78) -- (125.93,113.19) -- (139.31,113.19) -- cycle ;
\draw   (387.48,33.26) -- (401.36,33.26) -- (401.36,39.83) -- (387.48,39.83) -- cycle ;
\draw   (400.35,49.73) -- (397.64,54.23) -- (392.22,54.23) -- (389.51,49.73) -- (392.22,45.23) -- (397.64,45.23) -- cycle ;
\draw   (184,75) -- (177.31,86.59) -- (163.93,86.59) -- (157.24,75) -- (163.93,63.41) -- (177.31,63.41) -- cycle ;
\draw   (360.93,80.73) -- (354.24,92.32) -- (340.86,92.32) -- (334.17,80.73) -- (340.86,69.15) -- (354.24,69.15) -- cycle ;
\draw   (487.07,81.83) -- (480.38,93.42) -- (467,93.42) -- (460.31,81.83) -- (467,70.25) -- (480.38,70.25) -- cycle ;
\draw   (234.51,77.51) -- (248,91) -- (234.51,104.49) -- (221.02,91) -- cycle ;
\draw   (443.21,29.61) -- (449.37,35.51) -- (443.21,41.42) -- (437.05,35.51) -- cycle ;
\draw   (184,43.38) -- (177.31,54.97) -- (163.93,54.97) -- (157.24,43.38) -- (163.93,31.79) -- (177.31,31.79) -- cycle ;
\draw  [dash pattern={on 6pt off 0.75pt}][line width=0.75]  (384.76,26.71) -- (496.5,26.71) -- (496.5,59.6) -- (384.76,59.6) -- cycle ;
\draw  [dash pattern={on 3.75pt off 2.25pt}]  (248,91) -- (259.69,82.97) ;
\draw [shift={(261.33,81.83)}, rotate = 145.49] [color={rgb, 255:red, 0; green, 0; blue, 0 }  ][line width=0.75]    (8.74,-2.63) .. controls (5.56,-1.12) and (2.65,-0.24) .. (0,0) .. controls (2.65,0.24) and (5.56,1.12) .. (8.74,2.63)   ;
\draw  [dash pattern={on 3.75pt off 2.25pt}]  (218,59) -- (221.02,91) ;
\draw  [dash pattern={on 3.75pt off 2.25pt}]  (183.8,107.6) -- (221.02,91) ;
\draw  [dash pattern={on 3.75pt off 2.25pt}]  (360.93,80.73) -- (383,80.29) ;
\draw [shift={(385,80.25)}, rotate = 178.85] [color={rgb, 255:red, 0; green, 0; blue, 0 }  ][line width=0.75]    (8.74,-2.63) .. controls (5.56,-1.12) and (2.65,-0.24) .. (0,0) .. controls (2.65,0.24) and (5.56,1.12) .. (8.74,2.63)   ;
\draw    (113.6,75.1) -- (155.24,75) ;
\draw [shift={(157.24,75)}, rotate = 179.87] [color={rgb, 255:red, 0; green, 0; blue, 0 }  ][line width=0.75]    (8.74,-2.63) .. controls (5.56,-1.12) and (2.65,-0.24) .. (0,0) .. controls (2.65,0.24) and (5.56,1.12) .. (8.74,2.63)   ;
\draw    (312.13,80.83) -- (331.67,80.53) ;
\draw [shift={(333.67,80.5)}, rotate = 179.11] [color={rgb, 255:red, 0; green, 0; blue, 0 }  ][line width=0.75]    (8.74,-2.63) .. controls (5.56,-1.12) and (2.65,-0.24) .. (0,0) .. controls (2.65,0.24) and (5.56,1.12) .. (8.74,2.63)   ;
\draw    (107.5,117.8) -- (107.6,125.4) ;
\draw  [dash pattern={on 3.75pt off 2.25pt}]  (435.03,44.96) -- (448.7,44.65) ;
\draw [shift={(450.7,44.6)}, rotate = 178.68] [color={rgb, 255:red, 0; green, 0; blue, 0 }  ][line width=0.75]    (4.37,-1.32) .. controls (2.78,-0.56) and (1.32,-0.12) .. (0,0) .. controls (1.32,0.12) and (2.78,0.56) .. (4.37,1.32)   ;
\draw    (435.03,54.46) -- (448.7,54.15) ;
\draw [shift={(450.7,54.1)}, rotate = 178.68] [color={rgb, 255:red, 0; green, 0; blue, 0 }  ][line width=0.75]    (4.37,-1.32) .. controls (2.78,-0.56) and (1.32,-0.12) .. (0,0) .. controls (1.32,0.12) and (2.78,0.56) .. (4.37,1.32)   ;
\draw   (183.8,107.6) -- (177.11,119.19) -- (163.73,119.19) -- (157.04,107.6) -- (163.73,96.01) -- (177.11,96.01) -- cycle ;
\draw   (63.33,97.56) -- (112.4,97.56) -- (112.4,117.56) -- (63.33,117.56) -- cycle ;
\draw    (112.6,107.5) -- (155.04,107.6) ;
\draw [shift={(157.04,107.6)}, rotate = 180.13] [color={rgb, 255:red, 0; green, 0; blue, 0 }  ][line width=0.75]    (8.74,-2.63) .. controls (5.56,-1.12) and (2.65,-0.24) .. (0,0) .. controls (2.65,0.24) and (5.56,1.12) .. (8.74,2.63)   ;
\draw   (147,42.78) -- (140.31,54.37) -- (126.93,54.37) -- (120.24,42.78) -- (126.93,31.19) -- (140.31,31.19) -- cycle ;
\draw    (108.6,43.4) -- (118.24,42.89) ;
\draw [shift={(120.24,42.78)}, rotate = 176.95] [color={rgb, 255:red, 0; green, 0; blue, 0 }  ][line width=0.75]    (8.74,-2.63) .. controls (5.56,-1.12) and (2.65,-0.24) .. (0,0) .. controls (2.65,0.24) and (5.56,1.12) .. (8.74,2.63)   ;
\draw    (108.6,43.4) -- (108.49,65.36) ;
\draw   (204.51,45.51) -- (218,59) -- (204.51,72.49) -- (191.02,59) -- cycle ;
\draw  [dash pattern={on 3.75pt off 2.25pt}]  (184,43.38) -- (191.02,59) ;
\draw  [dash pattern={on 3.75pt off 2.25pt}]  (184,75) -- (191.02,59) ;
\draw    (438.47,82.67) -- (458,82.36) ;
\draw [shift={(460,82.33)}, rotate = 179.11] [color={rgb, 255:red, 0; green, 0; blue, 0 }  ][line width=0.75]    (8.74,-2.63) .. controls (5.56,-1.12) and (2.65,-0.24) .. (0,0) .. controls (2.65,0.24) and (5.56,1.12) .. (8.74,2.63)   ;
\draw    (107.33,125.17) -- (117.24,124.84) ;
\draw [shift={(119.24,124.78)}, rotate = 178.14] [color={rgb, 255:red, 0; green, 0; blue, 0 }  ][line width=0.75]    (8.74,-2.63) .. controls (5.56,-1.12) and (2.65,-0.24) .. (0,0) .. controls (2.65,0.24) and (5.56,1.12) .. (8.74,2.63)   ;

\draw (132.62,124.78) node  [font=\small]  {$e_{n}$};
\draw (170.62,75) node  [font=\small]  {$w_{m}$};
\draw (347.55,80.73) node  [font=\small]  {$s_{sup}$};
\draw (473.69,81.83) node  [font=\small]  {$d$};
\draw (234.51,91) node    {$\lor $};
\draw (170.62,43.38) node  [font=\small]  {$t_{os}$};
\draw (403.5,33.5) node [anchor=north west][inner sep=0.75pt]  [font=\tiny] [align=left] {events};
\draw (403.36,47.11) node [anchor=north west][inner sep=0.75pt]  [font=\tiny] [align=left] {fluents};
\draw (452.36,31) node [anchor=north west][inner sep=0.75pt]  [font=\tiny] [align=left] {operators};
\draw (452.36,41.5) node [anchor=north west][inner sep=0.75pt]  [font=\tiny] [align=left] {pre};
\draw (451.86,50) node [anchor=north west][inner sep=0.75pt]  [font=\tiny] [align=left] {eff};
\draw (88.2,107.56) node  [font=\small]  {$prod_{s}$};
\draw (287.87,81.63) node  [font=\small]  {$dis_{w}$};
\draw (412.6,81.93) node  [font=\small]  {$fau_{p}$};
\draw (170.42,107.6) node  [font=\normalsize]  {$w_{s}$};
\draw (91.4,75.16) node  [font=\small]  {$prod_{m}$};
\draw (133.62,42.78) node  [font=\small]  {$m_{k}$};
\draw (204.51,59) node    {$\land $};

\end{tikzpicture}}
    	\caption{Diagram of the $\kappa$ described in Example \ref{ex:formalisation}.}
    	\label{ex:formalisation_schema} 
    \end{figure}
    
    \begin{example}[\textit{pollution---formalisation}]\label{ex:formalisation}
        Figure \ref{ex:formalisation_schema} illustrates how Examples \ref{ex:preemption} and \ref{ex:duplication} can be formalised in this action language. These examples are composed of two actions: the production of medicines and the production of connected speakers which we denote $prod_m$ and $prod_s$ respectively. The first is formalised as having two effects, the availability of $k$ doses of medicine ($m_k \in \mathbb{F}$) and the existence of wastewater ($w_m \in \mathbb{F}$). The second also has two effects, employment for $n$ individuals ($e_n \in \mathbb{F}$) and the existence of wastewater ($w_s \in \mathbb{F}$). We add to these actions two exogenous events: the fact of discharging wastewater ($dis_w \in \mathbb{U}$) and the potable water plant fault ($fau_p \in \mathbb{U}$). The event $fau_p$ is simple, it is triggered when the pollution indicator of the lake is above the threshold ($s_{sup} \in \mathbb{F}$) and it has as effect the damage of the inhabitants who are deprived of drinking water ($d \in \mathbb{F}$). The event $dis_w$ is triggered either when there is wastewater from the speakers factory ($w_s$) or when there is wastewater from the medicine factory ($w_m$) and $w_m$ treatment plant is out of service ($t_{os} \in \mathbb{F}$). This event raises the pollution indicator of the lake above the threshold ($s_{sup}$).
            \begin{itemize}
                \item[] $pre(prod_m)=\top, eff(prod_m)=m_k\wedge w_m$
                \item[] $pre(prod_s)=\top, eff(prod_s)=e_n\wedge w_s$
                \item[] $pre(dis_w)=w_s\vee(w_m\wedge t_{os}), eff(dis_w)=s_{sup}$
                \item[] $pre(fau_p)=s_{sup}, eff(fau_p)=d$
            \end{itemize}
    \end{example}
    
    We now have a tool for time projection. Since our final objective is to evaluate the permissibility of actions in a given scenario, this is sufficient. However, given the flexibility of answer set programming in which we have translated our action language, we could move at low cost to a planning tool managing concurrency of events and exogenous events.
    
    \section{Adapted Causal Inquiry}
\label{sec:Causal}

    Of the many fields studying causality, our approach is especially close to \textit{tort law} whose interest is about causation in specific situations. Hence, works in this field are a good source of inspiration, with a large number of insights due to the still current discussions on the topic. In a series of influential papers \citep{wright_causation_1985,wright_causation_1988}, \citeauthor{wright_causation_1985} demonstrates how essential a causal inquiry is in the process of determining \textit{tort liability}. He emphasises the fundamental difference between causation and responsibility---or in the words of \citeauthor{vincent_structured_2011}'s taxonomy \citep{vincent_structured_2011}, between `causal responsibility' and `outcome responsibility'. In the process, he shows that this causal inquiry needs to be factual and independent of any policy choice. The argument we make is that such a causal inquiry is the same as the one that must be incorporated into proposals in computational ethics.
    
    Wright argues that a satisfying tort liability analysis---whose goal is to determine if a defendant is the `responsible cause' of an injury---requires a factual and independent of policy choices causal inquiry. In his papers Wright criticises the processes for determining responsibility for an injury in which the causal inquiry is flawed and polluted with subjective aspects---a process where causality and responsibility are conflated. \citeauthor{wright_causation_1985}'s initial observation is that those two notions are too often conflated \citep{wright_causation_1985}. The fact that `the phrase ``\textit{the cause}" is simply an elliptical way of saying ``\textit{the responsible cause}"' \citep{wright_causation_1985} shows how thin the boundary between those notions is. To clarify this conflation, \cite{wright_causation_1985} describes the process to determine if an individual is legally responsible for an injury. This process has three stages: (i) \textit{tortious-conduct inquiry}, where are identified the defendant's conducts that could potentially imply legal responsibility (intentional, negligent, hazardous, \dots); (ii) \textit{causal inquiry}, where is evaluated if the identified tortious conducts really contributed to cause the harm, i.e. if they can be considered as causes of the injury; (iii) \textit{proximate-cause inquiry}, where other causes of the injury are considered, so as to evaluate if they mitigate or eliminate the defendant's legal responsibility for the injury. Of those three stages, only the second is entirely factual and independent of policy choices. It determines if a conduct was a cause of the injury. The two others are subject to policy considerations that `determine which causes and consequences will give rise to liability' \citep{wright_causation_1985}. Not to yield into the easy confusion between responsibility and causality, our goal is to propose a definition of actual causality suitable for a causal inquiry as presented by Wright, i.e. factual and independent of policy choices.
    
    The actual causation definitions based solely on strong necessity---also known as counterfactual dependence---fail to capture the commonly accepted intuition on overdetermination cases (early preemption, late preemption, and symmetric overdetermination) \citep{clark_causation_2003,menzies_counterfactual_2020}. The commonly used in law \textit{But-for test}, also known as \textit{Conditio Sine Qua Non} \citep{satoh_disjunction_2006}, is one of those unsuccessful definitions---it is the basis of a significant part of the few works in computational ethics that integrates a mechanism to establish causal relationships \citep{lindner_hera_2017,lindner_formalization_2018}. This test states that `an act was a cause of an injury if and only if, but for the act, the injury would not have occurred' \citep{wright_causation_1985}. `In the context of structural equations, this flawed account can be described as equating causation with counterfactual dependence' \citep{beckers_counterfactual_2021}. Let us take Example \ref{ex:preemption} and apply the But-for test to it. Would the harm to the inhabitants of the village have occurred if the factory producing the speakers had not launched its production? Given the scenario the answer is yes, the harm would still have occurred due to the presence of $w_m$ discharges and the state $t_{os}$ of $w_m$ treatment plant. The production of speakers is therefore not a cause of the harm according to the But-for test because it is not necessary for the harm to occur. The same result is obtained if we apply it to the production of medicines. Hence, this test tells us that neither action is a cause of the harm---result which the reader will intuitively reject. Applying the But-for test to Example \ref{ex:duplication} gives exactly the same result. Given that overdetermination cases are not just hypothetical and rare cases (cases of pollution, suicide, economic loss, \dots), those strong necessity based approaches are not suitable for our purposes.
    
    The dominant approach of actual causality---HP definition \citep{halpern_actual_2016}---deals with those cases, but at the cost of the factualness of the causal inquiry. This definition has the same roots than the But-for test, \citeauthor{hume_enquete_1748}'s definition of causation second formulation \citep{hume_enquete_1748}:
    
        \begin{quote}
            [First formulation:] we may define a cause to be an object followed by another, and where all objects, similar to the first, are followed by objects similar to the second. [Second formulation:] Or, in other words, where if the first object had not been, the second had never existed.
        \end{quote}
    
    It is the result of an iterative process that originates in \citeauthor{pearl_causality_2000}'s formalisation \citep{pearl_causality_2000} of \citeauthor{lewis_causation_1973}'s vision \citep{lewis_causation_1973} in structural equations framework (SEF). HP approach is more complete than the But-for test in the sense that other elements in addition to counterfactual dependence where included in order to deal with some complex cases. One of those elements is interventionism. This assumption states that an event $C$ causes a second event $E$ if and only if, both events occur, and that, given an intervention allowing to fix the occurrence of a certain set of other events in the context---without being constrained to respect the physical coherence of the world---there is a context where if the first event had not occurred, the second would not have occurred either. This assumption is described by \cite{beckers_counterfactual_2021} using SEF notation as:
    
        \begin{quote}
            \textbf{Interventionism} They all share the assumption [HP-style definitions] that the relation between counterfactual dependence and causation takes on the following form: $C=c$ causes $E=e$ iff $E=e$ is counterfactually dependent on $C=c$ given an intervention $\vec{X} \leftarrow x$ that satisfies some conditions P. The divergence between these definitions is to be found in the condition P that should be satisfied.
        \end{quote}
    
    Interventionism---that \citeauthor{beckers_counterfactual_2021}'s CNESS \citep{beckers_counterfactual_2021} and \citeauthor{beckers_principled_2018}'s BV \citep{beckers_principled_2018} definitions reject---introduces non factual elements to the causal inquiry which appear problematic even for \cite{halpern_actual_2018}: 
    
        \begin{quote}
            if I fix BH [Billy hits] to zero here, I am sort of violating the way the world works. [...] I am contemplating counterfactuals are inconsistent with the equations but I seem to need to do that in order to get things to work out right. Believe me, we tried many other definitions.
        \end{quote}
    
    In addition to non factual elements, the divergence on which `conditions P' to apply can be equated with policy choices. These elements make HP-style definitions non adequate for our context.
    
    STIT approaches are also part of this family where strong necessity is central. Usual STIT approaches focus on the relationship between the agent and the states of the world. In order to be closer to the philosophical tradition according to which the actual causal relationship is defined between two events, we find action languages ideal because the events are central elements. Because of their modal approach, STIT works such as \citep{abarca_stit_2022,lorini_logical_2014} easily involve epistemic aspects---outside of the scope of this paper---fundamental when one wishes to go beyond causality by looking at responsibility.
    
    The NESS test which subordinates necessity to sufficiency is an approach that deals with overdetermination cases \citep{wright_causation_1985,wright_causation_1988,wright_ness_2011,baumgartner_regularity_2013} and that satisfies our inquiry needs. Introduced by \citeauthor{wright_causation_1985} in response to But-for test flaws, this test states that \citep{wright_causation_1985,wright_causation_1988}:
        
        \begin{quote}
            A particular condition was a cause of a specific consequence if and only if it was a necessary element of a set of antecedent actual conditions that was sufficient for the occurrence of the consequence.
        \end{quote}        
    
    Unlike approaches mentioned above, it belongs to a second high impact approach family, regularity theories of causation \citep{andreas_regularity_2021}. Those theories are also based on \citeauthor{hume_enquete_1748}'s definition of causation \citep{hume_enquete_1748}, but on the first formulation. Specifically, the NESS test is closer to \citeauthor{mill_system_1843}'s interpretation of this formulation \citep{mill_system_1843} which introduced that there are potentially a multiplicity of distinct, but equally sufficient sets of conditions. The NESS test is even closer to \citeauthor{mackie_cement_1980}'s proposal \citep{mackie_cement_1980}. Indeed, unlike \citeauthor{mill_system_1843}'s vision \citep{mill_system_1843} whereby the cause is the sufficient set, \cite{mackie_cement_1980} considers that each element of the set is a cause. 
    
    Examples \ref{ex:preemption} and \ref{ex:duplication} have both two potential sets of conditions sufficient to produce the inhabitants' harm, each set related to one of the possible actions. In Example \ref{ex:duplication}, the two sets have all their conditions met, and both the production of speakers and medicines are a necessary element for the occurrence of one of the two sets. Thus, the NESS test indicates that these two actions are causes of the harm---as intuitively expected. In Example \ref{ex:preemption}, only the set containing the production of speakers has all its conditions met. This action is therefore considered by the NESS test to be a cause of the harm, and we will say that it is a cause that preempts the effects of the production of medicines. The production of medicines will therefore not be considered a cause---as intuitively expected.
    
    The actual causation definition proposed in \citep{sarmiento_action_2022-1} is an action languages suitable formalisation of Wright's NESS test. Even if accepted by influential counterfactual theories of causation authors as embodying our basic intuition of causation---such as \citeauthor{pearl_causality_2000} \citep{beckers_counterfactual_2021}---criticism of the use of logic as formalism has prevented the popularisation of this test. What is argued is the inadequacy of logical sufficiency and logical necessity to formalise these intuitions. Recent works have shown that rejecting the formalism is not a reason to reject the idea behind it by successfully formalising the NESS test in causal calculus \citep{bochman_actual_2018} and in the structural equations framework \citep{beckers_counterfactual_2021}. It is conceivable to work on a way of compiling existing action languages problems and translating them into SEF. However, works have shown SEF flaws \citep{bochman_laws_2018} and that in complex evolving contexts \citep{halpern_axiomatizing_2000,hopkins_causality_2007,batusov_situation_2018} like ours, this translating approach is not necessarily desirable \citep{hopkins_causality_2007}:
    
        \begin{quote}
            Structural causal models are excellent tools for many types of causality-related questions. Nevertheless, their limited expressivity render them less than ideal for some of the more delicate causal queries, like actual causation. These queries require a language that is suited for dealing with complex, dynamically changing situations.
        \end{quote}
        
    Now that we have introduced the formalism in which we will represent the problems and the desired characteristics of the causal inquiry, all that remains to be done to link automated planning and causality---and computational ethics and causality by the same occasion---is to describe the action languages suitable definition of causation on which we rely, and then to introduce its translation into logic programming. Since we are in an operational framework, we can take some distance from metaphysical considerations and assume `that [causal laws] they are deterministic, and that they permit neither backwards causation nor causation across a temporal gap' \citep{clark_causation_2003}.

    \section{Actual Causality}
\label{sec:Actual_cause}
    
    In the context of action languages, we consider that a first event is an actual cause of a second event if and only if the occurrence of the first is a NESS-cause of the triggering of the second. As commonly accepted by philosophers, the relation of causality we aim to define links two events. However, `events are not the only things that can cause or be caused' \citep{lewis_causation_1973}. Action languages represent the evolution of the world as a succession of states produced by the occurrence of events, thus introducing states between events. Therefore, we need to define causal relations where causes are occurrence of events and effects are formulas of the language $\mathcal{P}$ truthfulness. This section will introduce definitions which establish such a relation based on \citeauthor{wright_causation_1985}'s NESS test of causation \citep{wright_causation_1985}.
    
    \begin{definition}[causal setting $\chi$]
        The action language causal setting denoted $\chi$ is the couple $(\sigma,\kappa)$ with $\sigma$ a scenario and $\kappa$ a context.
    \end{definition}
    
    From now on, when reference is made to events and states, they will be those from $\tau_{\sigma,\kappa}^e$ and $\tau_{\sigma,\kappa}^s$ respectively. Thus, the set of all events which actually occurred at time point $t$ is $E^{\chi}(t)=\tau_{\sigma,\kappa}^e(t)$. Following the same reasoning, the actual state at time point $t$ is $S^{\chi}(t)=\tau_{\sigma,\kappa}^s(t)$. For the sake of brevity, when a set of occurrences of events $C=\left\{(e,t),~e\in E^{\chi}(t),~t\in \mathbb{T}\right\}$ will be used in the context of the update operator $\triangleright$ or the predicate $actualEff(E,L)$, it will actually only refer to the events of the couples in this set\footnote{Meaning that $S\triangleright C \overset{\mathrm{def}}{=} S\triangleright\left\{e, \exists t, (e,t)\in C\right\}$ and $actualEff(C,S) \overset{\mathrm{def}}{=} actualEff(\left\{e, \exists t, (e,t)\in C\right\},S)$.}.
    
    \begin{definition}[direct NESS-causes]\label{def:direct_NESS}
        Given a causal setting $\chi$ and a couple $(\psi,t_{\psi})\in\mathcal{P}\times\mathbb{T}$ such that $S^{\chi}(t_{\psi})\vDash\psi$, the occurrence of events set $C=\left\{(e,t),~e\in E^{\chi}(t),~t\in \mathbb{T}\right\}$ is a sufficient set of direct NESS-causes of the truthfulness of the formula $\psi$ at $t_{\psi}$, denoted $C\rightsquigarrow(\psi,t_{\psi})$, iff there exists a partial state $W \subseteq Lit_{\mathbb{F}}$ that we call backing such that:
            \begin{itemize}
                \item[$\bullet$] Causal sufficiency and minimality of $W$: $W\vDash\psi$ and $\forall W'\subset W, ~W'\not\vDash\psi$.
                \item[] There is a decreasing sequence $t_1,\dots,t_k$ and a partition $W_1,\dots,W_k$ of $W$ such that $\forall i \in \left\{1,\dots,k\right\}$, given $C(t_i) = C \cap E^{\chi}(t_i)$:
                \begin{itemize}                
                    \item[$\bullet$] Weak necessity and minimality of $C$ at $t_i$: $S^{\chi}(t_i)\triangleright C(t_i)\vDash W_i$ and $\forall C'\subset C(t_i), ~S^{\chi}(t_i)\triangleright C'\not\vDash W_i$.
                    \item[$\bullet$] Persistency of necessity: $\forall t, ~t_i<t\leq t_\psi, ~S^{\chi}(t)\vDash W_i$.
                \end{itemize}
                \item[$\bullet$] Minimality of $C$: $C=\bigcup_{i\in\left\{1,\dots,k\right\}} C(t_i)$.
            \end{itemize}
        $(e,t)$ is a direct NESS-cause of $(\psi,t_\psi)$ iff $\exists C\subseteq\mathbb{E}\times\mathbb{T}$ such that $(e,t)\in C$ and $C\rightsquigarrow(\psi,t_{\psi})$. When relevant we make the backing $W$ explicit in the notation $C\underset{W}{\rightsquigarrow}(\psi,t_{\psi})$.
        \end{definition}
    
    \citeauthor{wright_causation_1985}'s NESS test \citep{wright_causation_1985} is based on three main principles which are formalised in Definition \ref{def:direct_NESS}: (i) sufficiency of a set, (ii) weak necessity of the conditions in that set, and (iii) actuality of the conditions. (i) In this definition, the sufficient set is the partial state $W$. More precisely, given the directionality embedded in Section \ref{sec:Modele} semantics, we have causal sufficiency that \cite{wright_ness_2011} differentiates from logical sufficiency: `The successional nature of causation is incorporated in the concept of causal sufficiency, which is defined as the complete instantiation of all the conditions in the antecedent of the relevant causal law'. Moreover, Definition \ref{def:direct_NESS} introduces the constraint of necessity and sufficiency minimality which has been proven to be essential for regularity theories of causation \citep{wright_ness_2011,baumgartner_regularity_2013,andreas_regularity_2021}. The minimality of $C$ condition ensures that weak necessity and minimality of C at $t_i$ is applied to all elements in the set. In other words, it excludes the possibility to have in $C$ an occurrence of event that has not occurred in one of the time points of the decreasing sequence $t_1,\dots,t_k$. (ii) Definition \ref{def:direct_NESS} formalises weak necessity by subordinating necessity to sufficiency achieving that \citep{wright_ness_2011}: `a causally relevant factor need merely be necessary for the sufficiency of a set of conditions sufficient for the occurrence of the consequence, rather than being necessary for the consequence itself'. It is worth mentioning that the condition $S^{\chi}(t_i)\not\vDash W_i$---intuitively expected when referring to necessity---is included in the minimality condition by the case where $C'= \varnothing$, thus $S^{\chi}(t_i) = S^{\chi}(t_i) \triangleright \varnothing \not\vDash W_i$. (iii) The actuality of the conditions is assured by the use of actual occurrence of events, which is implied by the presence of $E^{\chi}(t_i)$ in $C(t_i) = C\cap E^{\chi}(t_i)$. 
    
    Once causal sufficiency and minimality of the partial state $W$ is defined, the causal inquiry is conducted by a recursive reasoning on a partition of $W$. The goal of this recursive reasoning is to identify the events which occurrence was necessary to the sufficiency of $W$. This reasoning is done by going back in time and analysing the information given by $\tau_{\sigma,\kappa}^s(t)$ and $\tau_{\sigma,\kappa}^e(t)$. Two limit cases can be identified. The first is when the partition set $W_k$ is empty before its corresponding time $t_k$ is equivalent to $t=0$, meaning that all occurrences of events necessary for the sufficiency of $W$ have been identified. When this is not the case, it means that there are fluent literals in $W$ that were true in the initial state $S^{\chi}(0)$ and which value has not changed until $S^{\chi}(t_{\psi})$. In this second case, the set $C$ will contain the events $ini_l\in E^{\chi}(-1)$ whose $l$ remains in $W_k$---events which symbolise events in the past beyond the framework of formalisation.
    
    In practice, it is possible to study what will be considered as a direct NESS-causes of the truthfulness of $\psi$ a $t_{\psi}$ for each form that $\psi$ may take. In the case where $\psi$ is a fluent literal $l$, the direct NESS-causes will be the last occurrences of events to have made $l$ true before or at $t_{\psi}$. In this basic case $W$ is the singleton which unique element is that literal. This basic causal information is the one embedded in Section \ref{sec:Modele} action language semantics.
    
    \begin{proposition}[direct NESS-cause of a literal]\label{pro:direct_NESS_literal}
        Given a causal setting $\chi$, $e\in E^{\chi}(t)$, and a fluent literal $l\in Lit_{\mathbb{F}}$ true at $t_l$, $(e,t)$ is a direct NESS-cause of $(l,t_l)$ iff ($c_1$) $S^{\chi}(t)\triangleright \left\{e\right\}\vDash l$, ($c_2$) $\forall t',~t<t'\leq t_l$, $S^{\chi}(t')\vDash l$, and ($c_3$) $S^{\chi}(t)\not\vDash l$.
    \end{proposition}
    
    In the case where $\psi$ is a conjunction $\psi = l_1 \wedge \dots \wedge l_m$ of fluent literals, the direct NESS-causes will be all the occurrence of events that are direct NESS-causes of the truthfulness of one of the literals $l_j$ in the conjunction at $t_{\psi}$.
    
    \begin{proposition}[direct NESS-cause of a literal conjunction]\label{pro:direct_NESS_conjunction}
        Given a causal setting $\chi$, $e\in E^{\chi}(t)$, and a formula $\psi = l_1 \wedge \dots \wedge l_m$ of $\mathcal{P}$ true at $t_\psi$, $(e,t)$ is a direct NESS-cause of $(\psi,t_{\psi})$ iff ($c_1$) $\exists j\in \left\{1,\dots,m\right\},~S^{\chi}(t)\triangleright \left\{e\right\}\vDash l_j$, ($c_2$) $\forall t',~t<t'\leq t_\psi$, $S^{\chi}(t')\vDash l_j$, and ($c_3$) $S^{\chi}(t)\not\vDash l_j$.
    \end{proposition}
    
    Finally, the case where $\psi$ is in a disjunctive normal form is by far the more interesting and challenging. Indeed, it is in this case that we can be confronted to situations of overdetermination. Whenever $\psi$ is a disjunction, this means that there is a minimal causal sufficient backing $W$ for each disjunct. Each of these backings is a possible way to cause the truthfulness of the formula $\psi$ at $t_{\psi}$---in the same spirit as \citeauthor{beckers_counterfactual_2021}'s paths \citep{beckers_counterfactual_2021}.
    
    \begin{proposition}[direct NESS-cause of a DNF]\label{pro:direct_NESS_DNF}
        Given a causal setting $\chi$, $e\in E^{\chi}(t)$, and a DNF, subsumption minimal, and tautology free formula $\psi = \psi_1 \vee \dots \vee \psi_m$ of $\mathcal{P}$ such that $\psi\not\vDash\bot$, $(e,t)$ is a direct NESS-cause of $(\psi,t_{\psi})$ iff ($c_1$) $\exists j\in \left\{1,\dots,m\right\}$, $S^{\chi}(t_\psi)\vDash \psi_j$, ($c_2$) $S^{\chi}(t)\triangleright \left\{e\right\}\vDash l\in\psi_j$, ($c_3$) $\forall t',~t<t'\leq t_\psi$, $S^{\chi}(t')\vDash l$, and ($c_4$) $S^{\chi}(t)\not\vDash l$.
    \end{proposition}
    
    Example \ref{ex:switches} illustrates how Definition \ref{def:direct_NESS} handles one of those challenging situations and in which aspects it is different from \citep{batusov_situation_2018}.
    
        \begin{figure}[htp]
            \centering
            \begin{minipage}{.48\textwidth}
                \centering
                \vspace{0.9cm}
            	\resizebox{5cm}{!}{\tikzset{every picture/.style={line width=0.75pt}} 

\begin{tikzpicture}[x=0.75pt,y=0.75pt,yscale=-1,xscale=1]

\draw   (68.56,239.84) -- (96.21,239.84) (102.35,221) -- (102.35,258.68) (102.35,239.84) -- (130,239.84) (93.75,230.42) -- (96.21,230.42) -- (96.21,249.26) -- (93.75,249.26) -- (93.75,230.42) -- cycle ;
\draw   (47.17,139.96) -- (57.79,139.96) (89.65,139.96) -- (100.27,139.96) (62.04,138.95) -- (87.53,123.94) (85.4,139.96) .. controls (85.4,138.3) and (86.35,136.95) .. (87.53,136.95) .. controls (88.7,136.95) and (89.65,138.3) .. (89.65,139.96) .. controls (89.65,141.61) and (88.7,142.96) .. (87.53,142.96) .. controls (86.35,142.96) and (85.4,141.61) .. (85.4,139.96) -- cycle (57.79,139.96) .. controls (57.79,138.3) and (58.74,136.95) .. (59.91,136.95) .. controls (61.09,136.95) and (62.04,138.3) .. (62.04,139.96) .. controls (62.04,141.61) and (61.09,142.96) .. (59.91,142.96) .. controls (58.74,142.96) and (57.79,141.61) .. (57.79,139.96) -- cycle ;
\draw   (77.03,160.02) -- (87.65,160.02) (119.52,160.02) -- (130.14,160.02) (91.9,159.02) -- (117.39,144.01) (115.27,160.02) .. controls (115.27,158.36) and (116.22,157.02) .. (117.39,157.02) .. controls (118.57,157.02) and (119.52,158.36) .. (119.52,160.02) .. controls (119.52,161.68) and (118.57,163.02) .. (117.39,163.02) .. controls (116.22,163.02) and (115.27,161.68) .. (115.27,160.02) -- cycle (87.65,160.02) .. controls (87.65,158.36) and (88.61,157.02) .. (89.78,157.02) .. controls (90.95,157.02) and (91.9,158.36) .. (91.9,160.02) .. controls (91.9,161.68) and (90.95,163.02) .. (89.78,163.02) .. controls (88.61,163.02) and (87.65,161.68) .. (87.65,160.02) -- cycle ;
\draw   (107.17,180.16) -- (117.79,180.16) (149.65,180.16) -- (160.27,180.16) (122.04,179.15) -- (147.53,164.14) (145.4,180.16) .. controls (145.4,178.5) and (146.35,177.15) .. (147.53,177.15) .. controls (148.7,177.15) and (149.65,178.5) .. (149.65,180.16) .. controls (149.65,181.81) and (148.7,183.16) .. (147.53,183.16) .. controls (146.35,183.16) and (145.4,181.81) .. (145.4,180.16) -- cycle (117.79,180.16) .. controls (117.79,178.5) and (118.74,177.15) .. (119.91,177.15) .. controls (121.09,177.15) and (122.04,178.5) .. (122.04,180.16) .. controls (122.04,181.81) and (121.09,183.16) .. (119.91,183.16) .. controls (118.74,183.16) and (117.79,181.81) .. (117.79,180.16) -- cycle ;
\draw   (180.39,191.54) .. controls (189.35,191.69) and (196.51,198.47) .. (196.37,206.68) .. controls (196.24,214.9) and (188.86,221.43) .. (179.89,221.29) .. controls (170.93,221.14) and (163.77,214.36) .. (163.91,206.14) .. controls (164.04,197.93) and (171.42,191.39) .. (180.39,191.54) -- cycle (191.87,196.02) -- (168.41,216.81) (191.52,217.19) -- (168.76,195.63) (180.43,188.57) -- (180.39,191.54) (179.89,221.29) -- (179.84,224.26) ;
\draw    (179.84,224.26) -- (180,240) ;
\draw    (130,240) -- (180,240) ;
\draw    (180,160) -- (180.43,188.57) ;
\draw    (160,160) -- (180,160) ;
\draw    (160,140) -- (160,180) ;
\draw    (100,140) -- (160,140) ;
\draw    (130,160) -- (160,160) ;
\draw    (47.17,139.96) -- (47.17,179.96) ;
\draw    (47.17,179.96) -- (110,180) ;
\draw    (40,160) -- (47.17,159.96) ;
\draw    (40,160) -- (40,240) ;
\draw    (40,240) -- (68.56,239.84) ;
\draw    (47.17,159.96) -- (77.03,160.02) ;

\draw (152,199.9) node [anchor=north west][inner sep=0.75pt]  [font=\footnotesize]  {$\psi $};
\draw (80.25,222.9) node [anchor=north west][inner sep=0.75pt]  [font=\footnotesize]  {$l_{4}$};
\draw (129,179.9) node [anchor=north west][inner sep=0.75pt]  [font=\footnotesize]  {$l_{3}$};
\draw (98,160.4) node [anchor=north west][inner sep=0.75pt]  [font=\footnotesize]  {$l_{2}$};
\draw (68.25,139.9) node [anchor=north west][inner sep=0.75pt]  [font=\footnotesize]  {$l_{1}$};

\end{tikzpicture}}
            	\caption{Electrical circuit consisting of a voltage source, three switches, and an individual connected to electrodes.}
            	\label{ex:switches_schema}
            \end{minipage}
            \hspace{0.2cm}
            \begin{minipage}{.48\textwidth}
                \centering
            	\resizebox{6.5cm}{!}{\tikzset{every picture/.style={line width=0.75pt}} 

\begin{tikzpicture}[x=0.75pt,y=0.75pt,yscale=-1,xscale=1]

\draw [line width=0.75]    (30.59,269.16) -- (287.64,268.75) ;
\draw [shift={(289.64,268.75)}, rotate = 179.91] [color={rgb, 255:red, 0; green, 0; blue, 0 }  ][line width=0.75]    (10.93,-3.29) .. controls (6.95,-1.4) and (3.31,-0.3) .. (0,0) .. controls (3.31,0.3) and (6.95,1.4) .. (10.93,3.29)   ;
\draw [line width=0.75]    (30.59,269.16) -- (29.55,110.07) ;
\draw [line width=0.75]    (111.09,269.5) -- (110.55,109.93) ;
\draw [line width=0.75]    (190.76,269.67) -- (190.48,110.23) ;
\draw [line width=0.75]    (270.59,268.65) -- (270.15,110.29) ;
\draw [line width=0.75]  [dash pattern={on 4.5pt off 4.5pt}]  (85.59,269.06) -- (85.05,110.07) ;
\draw [line width=0.75]  [dash pattern={on 4.5pt off 4.5pt}]  (166.09,270.12) -- (165.48,109.89) ;
\draw [line width=0.75]  [dash pattern={on 4.5pt off 4.5pt}]  (245.59,268.74) -- (245.48,109.71) ;
\draw [line width=0.75]    (245.54,160.23) -- (190.62,159.95) ;
\draw [line width=0.75]    (245.22,190.16) -- (192.76,190.33) ;
\draw [shift={(192.76,190.33)}, rotate = 359.81] [color={rgb, 255:red, 0; green, 0; blue, 0 }  ][line width=0.75]    (0,5.59) -- (0,-5.59)   ;
\draw [line width=0.75]    (165.79,160.01) -- (112.62,159.69) ;
\draw [shift={(112.62,159.69)}, rotate = 0.35] [color={rgb, 255:red, 0; green, 0; blue, 0 }  ][line width=0.75]    (0,5.59) -- (0,-5.59)   ;
\draw [line width=0.75]    (82.51,130.35) -- (31.62,130.13) ;
\draw [shift={(31.62,130.13)}, rotate = 0.25] [color={rgb, 255:red, 0; green, 0; blue, 0 }  ][line width=0.75]    (0,5.59) -- (0,-5.59)   ;
\draw [shift={(82.51,130.35)}, rotate = 0.25] [color={rgb, 255:red, 0; green, 0; blue, 0 }  ][line width=0.75]    (0,5.59) -- (0,-5.59)   ;
\draw [line width=0.75]    (245.54,220.45) -- (192.39,219.79) ;
\draw [shift={(192.39,219.79)}, rotate = 0.71] [color={rgb, 255:red, 0; green, 0; blue, 0 }  ][line width=0.75]    (0,5.59) -- (0,-5.59)   ;
\draw [line width=0.75]    (245.47,251.18) -- (192.33,250.52) ;
\draw [shift={(192.33,250.52)}, rotate = 0.71] [color={rgb, 255:red, 0; green, 0; blue, 0 }  ][line width=0.75]    (0,5.59) -- (0,-5.59)   ;

\draw (283,271.99) node [anchor=north west][inner sep=0.75pt]  [font=\footnotesize]  {$t$};
\draw (65.5,271.4) node [anchor=north west][inner sep=0.75pt]  [font=\footnotesize]  {$0$};
\draw (145.5,272.4) node [anchor=north west][inner sep=0.75pt]  [font=\footnotesize]  {$1$};
\draw (225.5,271.4) node [anchor=north west][inner sep=0.75pt]  [font=\footnotesize]  {$2$};
\draw (14.5,242.41) node [anchor=north west][inner sep=0.75pt]  [font=\footnotesize]  {$\psi $};
\draw (14,212.91) node [anchor=north west][inner sep=0.75pt]  [font=\footnotesize]  {$l_{4}$};
\draw (14.5,182.91) node [anchor=north west][inner sep=0.75pt]  [font=\footnotesize]  {$l_{3}$};
\draw (14,152.91) node [anchor=north west][inner sep=0.75pt]  [font=\footnotesize]  {$l_{2}$};
\draw (14,122.91) node [anchor=north west][inner sep=0.75pt]  [font=\footnotesize]  {$l_{1}$};
\draw (84,101.9) node [anchor=north west][inner sep=0.75pt]  [font=\tiny]  {$E^{\chi }( 0)$};
\draw (164,102.04) node [anchor=north west][inner sep=0.75pt]  [font=\tiny]  {$E^{\chi }( 1)$};
\draw (243.5,102.04) node [anchor=north west][inner sep=0.75pt]  [font=\tiny]  {$E^{\chi }( 2)$};
\draw (91.5,151.91) node [anchor=north west][inner sep=0.75pt]  [font=\footnotesize]  {$e_{2}$};
\draw (173,182.91) node [anchor=north west][inner sep=0.75pt]  [font=\footnotesize]  {$e_{3}$};
\draw (88.5,123.41) node [anchor=north west][inner sep=0.75pt]  [font=\footnotesize]  {$e_{\neg 1}$};
\draw (172.2,212.91) node [anchor=north west][inner sep=0.75pt]  [font=\footnotesize]  {$e_{4}$};

\end{tikzpicture}}
            	\caption{Example \ref{ex:switches} evolution of fluents given $\kappa$.}
            	\label{ex:switches_sequence} 
            \end{minipage}
        \end{figure}
        
    \begin{example}[parallel switches and Milgram]\label{ex:switches}
        Consider Figure \ref{ex:switches_schema} simple electric circuit inspired by \citeauthor{milgram_behavioral_1963}'s experiment \citep{milgram_behavioral_1963}. This circuit is made up of a voltage source, an individual strapped and connected to electrodes, and three switches connected in parallel. The positive literals $l_1,l_2,l_3,l_4\in Lit_{\mathbb{F}}$ represent the closed state of each switch and the voltage source respectively---their respective complement thus represents the opened state. $\psi=(l_1\wedge l_4)\vee(l_2\wedge l_4)\vee(l_3\wedge l_4)$ where $\psi\in\mathcal{P}$ represents the triggering conditions for the strapped individual being electrocuted. Thus, three backings are possible to cause $\psi$: $W=\left\{l_1,l_4\right\}$, $W'=\left\{l_2,l_4\right\}$, and $W''=\left\{l_3,l_4\right\}$. $e_1,e_2,e_3\in\mathbb{E}$ are the events which intrinsic effect is to close each switch respectively, $e_4\in\mathbb{E}$ is an event which intrinsic effect is to close the voltage source, and $e_{\neg1}\in\mathbb{E}$ is the event which intrinsic effect is to open the first switch. We assume that the situation involves five agents: the one strapped and four others---each controlling one of the four components of the circuit. The studied sequences illustrated by Figure \ref{ex:switches_sequence} and given by $\tau_{\sigma,\kappa}^e$ and $\tau_{\sigma,\kappa}^s$ are:
            \begin{itemize}
                \item[] $E^{\chi}(-1)=\left\{ini_{l_1},ini_{\overline{l_2}},ini_{\overline{l_3}},ini_{\overline{l_4}}\right\}$ 
                \item[] $S^{\chi}(0)=\left\{l_1,\overline{l_2},\overline{l_3},\overline{l_4}\right\}, E^{\chi}(0)=\left\{e_{\neg 1},e_2\right\}$
                \item[] $S^{\chi}(1)=\left\{\overline{l_1},l_2,\overline{l_3},\overline{l_4}\right\}, E^{\chi}(1)=\left\{e_3,e_4\right\}$ 
                \item[] $S^{\chi}(2)=\left\{\overline{l_1},l_2,l_3,l_4\right\}$
            \end{itemize}
        Given the above traces, $\psi$ is true at $t=2$ by both $W'$ and $W''$.
    \end{example}
    
    The question that arises in Example \ref{ex:switches} is: what are the causes of $\psi$ being true at $t=2$? Said in another way, what are the causes of the strapped individual being electrocuted at $t=2$? \citeauthor{batusov_situation_2018}'s proposal \citep{batusov_situation_2018} will consider $(ini_{l_1},-1)$ and $(e_4,1)$ as `achievement causes', and $(e_2,0)$ as a `maintenance cause'---`causes responsible for protecting a previously achieved effect, despite potential threats that could destroy the effect' \citep{batusov_situation_2018}---this given that we omit to consider $(e_3,1)$ in the comparison because it occurs at the same time as $(e_4,1)$ and thus requires definitions that handle concurrency. Considering factuality as an essential feature of a causal inquiry, the presence of $(ini_{l_1},-1)$ in the causes seems unacceptable. Factually, $(ini_{l_1},-1)$ plays no role in the truthfulness of $\psi$ at $t=2$. Definition \ref{def:direct_NESS} gives us the direct NESS-cause relations $C'\underset{W'}{\rightsquigarrow}(\psi,2)$ and $C''\underset{W''}{\rightsquigarrow}(\psi,2)$, where $C'=\left\{(e_2,0),(e_4,1)\right\}$ and $C''=\left\{(e_3,1),(e_4,1)\right\}$ which union gives the answer $\left\{(e_2,0),(e_3,1),(e_4,1)\right\}$.
    
    The interpretation given by \cite{batusov_situation_2018} of Example \ref{ex:switches} is not the only possible divergent interpretation. We wondered whether answer $\left\{(e_2,0),(e_4,1)\right\}$ alone was not more satisfactory given that, even if both $l_2$ and $l_3$ are true at $t=2$, the precedence of $l_2$ could be taken into account. However, this intuition appears as conflating causality and responsibility. If we strictly limit ourselves to a factual causal inquiry as prescribed by \cite{wright_causation_1985}, both $(e_2,0)$ and $(e_3,1)$ are causes of the truthfulness of $\psi$ at $t=2$. The intuition that would induce us to take into account the precedence of $(e_2,0)$ belongs to \citeauthor{wright_causation_1985}'s \textit{proximate-cause inquiry} \citep{wright_causation_1985} and not to the causal inquiry. Indeed, once $(e_2,0)$ and $(e_3,1)$ are identified as causes, there is a policy choice for which the precedence of $(e_2,0)$ mitigates or eliminates the responsibility of $(e_3,1)$ for the final effect. We suspect that \citeauthor{batusov_situation_2018}'s choices \citep{batusov_situation_2018}---which would lead them to consider $(ini_{l_1},-1)$ as a cause---were influenced by this same intuition, but taken even further.
    
    \begin{proposition}[non unicity of direct NESS-causes]\label{pro:non_unicity}
        Given a causal setting $\chi$ and the causal relation $C\underset{W}{\rightsquigarrow}(\psi,t_{\psi})$, $C$ is not the unique set possibly satisfying $C\underset{W}{\rightsquigarrow}(\psi,t_{\psi})$.
    \end{proposition}
    
    \begin{proof}
        Consider $l\in Lit_\mathbb{F}$, the formula $\psi=l$, the sets of occurrence of events $C=\left\{(e,0)\right\}$ and $C'=\left\{(e',0)\right\}$ where $eff(e)=eff(e')=l$, and simple sequences given by $\tau_{\sigma,\kappa}^e$ and $\tau_{\sigma,\kappa}^s$ :
            \begin{itemize}
                \item[] $E^{\chi}(-1)=\left\{ini_{\overline{l}}\right\}$ 
                \item[] $S^{\chi}(0)=\left\{\overline{l}\right\}, E^{\chi}(0)=\left\{e,e'\right\}$
                \item[] $S^{\chi}(1)=\left\{l\right\}$
            \end{itemize}
        Consider an hypothetical backing $W=\left\{l\right\}$ of the truthfulness of $l$ at time $t=1$. This backing $W$ satisfies the causally sufficient and minimality conditions since $\left\{l\right\}\vDash l$ and that its only subset---namely $\varnothing$---does not satisfy $l$. There is a decreasing sequence $t_1=0$ and a partition $W_1=\left\{l\right\}$ of $W$ such that $\forall i\in\left\{1\right\}$, given $C(0) = C \cap E^{\chi}(0)=\left\{(e,0)\right\}$: $S^{\chi}(0)\triangleright C(0)\vDash \left\{l\right\}$---that is weak necessity of $C$ at $t_1$---$\forall C''\subset C(0), ~S^{\chi}(0)\triangleright C''\not\vDash \left\{l\right\}$---that is minimality of $C$ at $t_1$---and $\forall t, ~0<t\leq 1, ~S^{\chi}(t)\vDash \left\{l\right\}$---that is persistency of necessity. Finally, it is straightforward that $C=\bigcup_{i\in\left\{1\right\}} C(t_i)$---which is minimality of $C$. Hence, according to Definition \ref{def:direct_NESS}, $C\underset{W}{\rightsquigarrow}(\psi,t_{\psi})$. By applying the same reasoning with $C'(0) = C' \cap E^{\chi}(0)=\left\{(e',0)\right\}$ we obtain $C'\underset{W}{\rightsquigarrow}(\psi,t_{\psi})$. We have therefore proven non unicity of direct NESS-causes, even when taking the same backing $W$. This result is possible because we allow concurrency of events.
    \end{proof}
    
    Definition \ref{def:direct_NESS} gives us essential information about causal relations by looking to the actual effects of events. However, the set of direct NESS-causes of an effect may include exogenous events that are not necessarily relevant. This is especially true in a framework such as ours, where we are interested in the ethical dimension of an agent's decisions---thus actions. It is therefore essential to establish a causal chain by going back in time in order to find the set of actions that led to the effect. To this end, we must broaden our vision to look not only at the actual effects of events which are direct NESS-causes, but also at the events that caused those events to be triggered. We thus introduce NESS-causes that are found relying on the establishment of the causal chain.
    
    \begin{definition}[NESS-causes]\label{def:NESS}
        Given a causal setting $\chi$, the direct NESS-cause relation $C\underset{W}{\rightsquigarrow}(\psi,t_{\psi})$, and the decreasing sequence $t_1,\dots,t_k$ induced by the existing partition $W_1,\dots,W_k$ of the backing $W$, the occurrence of events set $C'=\left\{(e,t),~e\in E^{\chi}(t),~t\in \mathbb{T}\right\}$ is a sufficient set of NESS-causes of the truthfulness of the formula $\psi$ at $t_{\psi}$ iff one of the following cases is satisfied:
            \begin{itemize}
                \item Base case: $C'=C$.
                \item Recursive case: Given the set $C_R=C\setminus C'$ of `removable' occurrence of events and the partitions of $C$ and $C_R$ matching the decreasing sequence $t_1,\dots,t_k$---$C(t_1),\dots,C(t_k)$ and $C_R(t_1),\dots,C_R(t_k)$ respectively---there is a sequence of subsets $C_{O_{1}},\dots,C_{O_{k}}$---not necessarily monotonic in time---of the set of `overwhelming' occurrence of events $C_O=C'\setminus C$ such that:
                    \begin{itemize}
                        \item $C_O=\bigcup_{i\in \left\{1,\dots,k\right\}}C_{O_{i}}$.
                        \item $\forall i\in \left\{1,\dots,k\right\}, C_R(t_i)=\varnothing\implies C_{O_{i}}=\varnothing$.
                        \item $\forall i\in \left\{1,\dots,k\right\}, C_R(t_i)\neq\varnothing\implies C_{O_{i}}$ is a sufficient set of NESS-causes of $(tri(C_R(t_i)),t_{i})$.
                    \end{itemize}
            \end{itemize}
        $(e,t)$ is a NESS-cause of $(\psi,t_\psi)$ iff $\exists C'\subseteq\mathbb{E}\times\mathbb{T}$ such that $(e,t)\in C'$ and the occurrence of events set $C'$ is a sufficient set of NESS-causes of $(\psi,t_\psi)$. The set of NESS-causes $D=C\setminus C_R\cup C'$ is called a set of decisional causes if $D\subseteq\mathbb{A}\times\mathbb{T}$.
    \end{definition}
    
    Definition \ref{def:NESS} captures the way in which the occurrence of an event can have a causal relation---other than being a direct NESS-cause---with the truthfulness of $\psi$ at $t_\psi$: by being a NESS-cause of the triggering conditions of an occurrence of event that is a NESS-cause of $(\psi,t_\psi)$---captured by the conjunct $tri(C_R(t_i))$.
    
    \begin{example}[reconstituting the causal chain]\label{ex:ness}
        Consider the formalisation of Example \ref{ex:formalisation} and the sequence corresponding to Example \ref{ex:duplication} given by $\tau_{\sigma,\kappa}^e$ and $\tau_{\sigma,\kappa}^s$:
            \begin{itemize}
                \item[] $E^{\chi}(-1)=\left\{ini_{t_{os}},ini_{\neg m_{k}},ini_{\neg w_{m}},ini_{\neg w_{s}},ini_{\neg e_{n}},ini_{\neg s_{sup}},ini_{\neg d}\right\}$ 
                \item[] $S^{\chi}(0)=\left\{t_{os},\neg m_{k},\neg w_{m},\neg w_{s},\neg e_{n},\neg s_{sup},\neg d\right\}, E^{\chi}(0)=\left\{prod_m,prod_s\right\}$
                \item[] $S^{\chi}(1)=\left\{t_{os},m_{k},w_{m},w_{s},e_{n},\neg s_{sup},\neg d\right\}, E^{\chi}(1)=\left\{dis_w\right\}$ 
                \item[] $S^{\chi}(2)=\left\{t_{os},m_{k},w_{m},w_{s},e_{n},s_{sup},\neg d\right\}, E^{\chi}(2)=\left\{fau_p\right\}$ 
                \item[] $S^{\chi}(3)=\left\{t_{os},m_{k},w_{m},w_{s},e_{n},s_{sup},d\right\}$
            \end{itemize}
        Given the above sequence, Definition \ref{def:NESS} gives us as sufficient sets of NESS-causes of $(d,3)$: $\left\{(fau_p,2)\right\}$---corresponding to base case---$\left\{(dis_w,1)\right\}$, and $\left\{(prod_m,0),(prod_s,0),(ini_{t_{os}},-1)\right\}$, the last being the closest to a decisional set that can be obtained here.
    \end{example}
        
        \begin{figure}[htp]
            \centering
            \begin{minipage}{.48\textwidth}
                \centering
            	\resizebox{8cm}{!}{\input{ness_example}}
            	\caption{Some Example \ref{ex:duplication} causal relations.}
        	    \label{ex:causal_chain}
            \end{minipage}
            \hspace{0.3cm}
            \begin{minipage}{.48\textwidth}
                \centering
            	\resizebox{5.8cm}{!}{\tikzset{every picture/.style={line width=0.75pt}} 

\begin{tikzpicture}[x=0.75pt,y=0.75pt,yscale=-1,xscale=1]

\draw  [fill={rgb, 255:red, 74; green, 74; blue, 74 }  ,fill opacity=0.69 ] (51,112.08) .. controls (51,74.01) and (91.32,43.15) .. (141.05,43.15) .. controls (190.78,43.15) and (231.1,74.01) .. (231.1,112.08) .. controls (231.1,150.14) and (190.78,181) .. (141.05,181) .. controls (91.32,181) and (51,150.14) .. (51,112.08) -- cycle ;
\draw  [fill={rgb, 255:red, 155; green, 155; blue, 155 }  ,fill opacity=0.43 ] (161.02,82.56) .. controls (153.72,54.19) and (189.32,20.5) .. (240.53,7.32) .. controls (291.74,-5.85) and (339.18,6.46) .. (346.48,34.84) .. controls (353.78,63.21) and (318.18,96.9) .. (266.97,110.08) .. controls (215.76,123.25) and (168.32,110.94) .. (161.02,82.56) -- cycle ;
\draw  [fill={rgb, 255:red, 155; green, 155; blue, 155 }  ,fill opacity=0.43 ] (159.74,143.42) .. controls (170.53,116.18) and (219.14,109.9) .. (268.3,129.38) .. controls (317.46,148.87) and (348.56,186.74) .. (337.76,213.98) .. controls (326.97,241.22) and (278.36,247.5) .. (229.2,228.02) .. controls (180.04,208.53) and (148.94,170.66) .. (159.74,143.42) -- cycle ;

\draw (58.9,103) node [anchor=north west][inner sep=0.75pt]    {$( fau_{p} ,2)$};
\draw (221.1,205.98) node [anchor=north west][inner sep=0.75pt]    {$( ini_{t_{os}} ,-1)$};
\draw (237,44.4) node [anchor=north west][inner sep=0.75pt]    {$( dis_{w} ,1)$};
\draw (222.5,161.4) node [anchor=north west][inner sep=0.75pt]    {$( prod_{s} ,0)$};
\draw (220.5,184.9) node [anchor=north west][inner sep=0.75pt]    {$( prod_{m} ,0)$};
\draw (63.5,43.9) node [anchor=north west][inner sep=0.75pt]    {$C$};
\draw (313,90.4) node [anchor=north west][inner sep=0.75pt]    {$C_{1} '$};
\draw (114.5,47.9) node [anchor=north west][inner sep=0.75pt]  [font=\footnotesize]  {$C_{R}$};
\draw (277,5.9) node [anchor=north west][inner sep=0.75pt]  [font=\footnotesize]  {$C_{O}$};
\draw (161,62.9) node [anchor=north west][inner sep=0.75pt]  [font=\footnotesize]  {$C\cap C_{1} '$};
\draw (313,140.9) node [anchor=north west][inner sep=0.75pt]    {$C_{2} '$};
\draw (275,140.9) node [anchor=north west][inner sep=0.75pt]  [font=\footnotesize]  {$C_{O}$};
\draw (174.5,124.9) node [anchor=north west][inner sep=0.75pt]  [font=\footnotesize]  {$C\cap C_{2} '$};

\end{tikzpicture}}
            	\vspace{-0.3cm}
            	\caption{Example \ref{ex:duplication} sufficient sets of direct NESS-causes ($C$) and NESS-causes ($C$,$C_1'$,$C_2'$) of $(d,3)$.}
            	\label{ex:ness_sets} 
            \end{minipage}
        \end{figure}
    
    The question that arises is: how do we get the set $\left\{(prod_m,0),(prod_s,0),(ini_{t_{os}},-1)\right\}$ of NESS-causes of $(d,3)$---corresponding to $C_2'$ in Figure \ref{ex:ness_sets}---mentioned in Example \ref{ex:ness}? In this example, the partition of the backing $W=\left\{d\right\}$ in $C\underset{W}{\rightsquigarrow}(d,3)$ is $W_1=\left\{d\right\}$. The corresponding time sequence is $t_1=2$, the partition of the occurrence of events set $C$ is $C(t_1)=\left\{(fau_p,2)\right\}$, and the partition of the `removable' occurrence of events set $C_R$ is $C_R(t_1)=\left\{(fau_p,2)\right\}$. None of the occurrence of events in $C_R$ being the occurrence of an action implying an agent's volition which ends the causal chain---at least `strong' \citep{berreby_event-based_2018} causal relations as the ones defined by Definitions \ref{def:direct_NESS}, \ref{def:NESS}, and \ref{def:actual}---all the occurrences in $C_R$ are `removable'. The importance of different strength causal relations appears in many real cases as in bioethics with the difference between euthanasia and assisted suicide. According to Definition \ref{def:NESS} recursive case, $C_{O_{1}}$ is a sufficient set of NESS-causes of $(tri(fau_p),2)$. Given that $tri(fau_p)=s_{sup}$, we are looking for a sufficient set of NESS-causes of $(s_{sup},2)$. According to Definition \ref{def:direct_NESS}, $C''=\left\{(dis_w,1)\right\}$ is a sufficient set of direct NESS-cause of $(s_{sup},2)$. Hence, according to Definition \ref{def:NESS} base case, $C''=C_1'=\left\{(dis_w,1)\right\}$ is a sufficient set of NESS-causes of $(s_{sup},2)$. Therefore, according to Definition \ref{def:NESS} recursive case, $C_{O_{1}}=C_1'$ and thus $C_1'$ is a sufficient set of NESS-causes of $(d,3)$.
    
    However, $C_1'$ is not the only sufficient set of NESS-causes of $(s_{sup},2)$, it is only the one corresponding to the base case of Definition \ref{def:NESS}. We are thus going to apply the same reasoning that above but this time by considering the direct NESS-cause relation $C_1'\underset{W'}{\rightsquigarrow}(s_{sup},2)$. The partition of the backing $W'=\left\{s_{sup}\right\}$ is $W_1'=\left\{s_{sup}\right\}$. The corresponding time sequence is $t_1'=1$, the partition of the occurrence of events set $C_1'$ is $C_1'(t_1')=\left\{(dis_w,1)\right\}$, and the partition of the `removable' occurrence of events set $C_R'$ is $C_R'(t_1')=\left\{(dis_w,1)\right\}$. As above, all the occurrences in $C_R'$ are `removable'. According to Definition \ref{def:NESS}, $C_{O_{1}}'$ is a sufficient set of NESS-causes of $(tri(dis_w),1)$. Given that $tri(dis_w)=w_s\vee (w_m\wedge t_{os})$, we are looking for a sufficient set of NESS-causes of $(w_s\vee (w_m\wedge t_{os}),1)$. According to Definition \ref{def:direct_NESS}, $C'''=\left\{(prod_m,0),(prod_s,0),(ini_{t_{os}},-1)\right\}$ is a sufficient set of direct NESS-cause of $(w_s\vee (w_m\wedge t_{os}),1)$. Hence, according to Definition \ref{def:NESS} base case, $C'''=C_2'=\left\{(prod_m,0),(prod_s,0),(ini_{t_{os}},-1)\right\}$ is a sufficient set of NESS-causes of $(s_{sup},2)$. Therefore, according to Definition \ref{def:NESS} recursive case, $C_{O_{1}}=C_2'$ and thus $C_2'$ is also a sufficient set of NESS-causes of $(d,3)$.
    
    According to Definition \ref{def:NESS}, there is no other sufficient set of NESS-causes of $(d,3)$. Indeed, if we wanted to iterate the process we would be faced with the fact that neither $(ini_{t_{os}},-1)$, $(prod_m,0)$, or $(prod_s,0)$ can be removed and replaced. The first because of our bounded past formalisation. The second and third because it is the occurrence of an action implying an agent's volition which ends the causal chain. Therefore---according to Definition \ref{def:NESS} recursive case---given that $C_R=\varnothing$, $C_{O}=\varnothing$.
 
    Having determined the causal relations linking events and formulas of the language, we can now give a suitable for action languages definition of actual causality.
    
    \begin{definition}[actual cause]\label{def:actual}
        Given a causal setting $\chi$ and an event $e\in E^{\chi}(t_{\psi})$, the actual causes of $(e,t_{\psi})$ are the NESS-causes of $(tri(e),t_\psi)$, i.e. the truthfulness of the triggering conditions of $e$ at $t_{\psi}$.
    \end{definition}
    
\begin{figure}[htp]
        \centering
    	\resizebox{9.5cm}{!}{\input{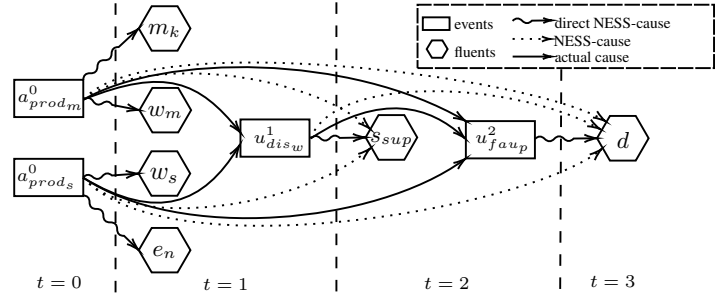}}
    	\caption{Causal relations in Example \ref{ex:duplication}.}
    	\label{ex:formalisation_res} 
    \end{figure}

    \begin{example}[\textit{pollution---causal relations}]\label{ex:causal_results}
        Figure \ref{ex:formalisation_res} illustrates Example \ref{ex:duplication} sequence of events as well as the causal relations that can be established based on the proposed definitions. We can see that as expected, both the launching of the production of connected speakers $(prod_s,0)$ and medicines $(prod_m,0)$ are actual causes of the potable water plant fault $(fau_p,2)$ and are also NESS-causes of the harm $(d,3)$. Given the sequence of events in Example \ref{ex:preemption}, the only action that would be identified as a cause of the harm is the launching of the production of connected speakers $(prod_s,0)$---as expected. Indeed, $(prod_m,1)$ is an actual cause of $(dis_w,2)$, but given that $s_{sup}\in S^{\chi}(2)$, $s_{sup}\not\in actualEff(dis_w,S^{\chi}(2))$. Hence, $\forall t\in\mathbb{T}$ there is no causal chain between $(prod_m,1)$ and $(fau_p,t)$.
    \end{example}
    
    \section{Logic Program for Actual Causality}
\label{sec:Logic_program}
    
    \cite{powers_ethics_2020} identify several challenges that need to be addressed for successfully modeling ethical reasoning. \citeauthor{sarmiento_action_2022-1}'s proposal of an actual causality definition adapted to action languages \citep{sarmiento_action_2022-1} contributes to tackle one of those challenges: `to evaluate the consequences of actions'. We chose to use answer set programming (ASP) for the implementation of this proposal. This choice obeys our will to tackle another challenge mentioned: be able to `overcome logical contradictions' that arise commonly in ethical reasoning because of conflicts of norms. Indeed, ASP is a form of non-monotonic declarative programming that has shown its capabilities in dealing with ethical issues \citep{ganascia_non-monotonic_2015}.
    
    In ASP, problems are encoded as extended disjunctive programs \citep{gelfond_classical_1991}. Those are finite logic theories from which are extracted answer sets which are stable models. An extended disjunctive program is a set of rules $r$ of the form: $L_1;\dots;L_l$ $\leftarrow$ $L_{l+1},\dots,L_m,$ $not~L_{m+1},\dots,not~L_n$ where each $L_i\in Lit$ is a positive or negative literal---respectively $A$ and $\neg A$ for an atom $A$---$not$ is negation as failure, symbol `$;$' represents disjunction, symbol `$,$' represents conjunction, and $n\geq m\geq l\geq 0$. We denote $head(r)=\left\{L_1,\dots,L_l\right\}$, $body^{+}(r)=\left\{L_{l+1},\dots,L_m\right\}$, and $body^{-}(r)=\left\{L_{m+1},\dots,L_n\right\}$. Rules are considered integrity constraints if $head(r)=\varnothing$ and facts if $body(r)=\varnothing$. A program $\Pi$ with variables is semantically identified with its ground instantiation. The semantics of extended disjunctive programs is given by the answer set semantics \citep{gelfond_classical_1991}. A set $S\subseteq Lit$ satisfies a rule $r$ if $body^{+}(r)\subseteq S$ and $body^{-}(r)\cap S=\varnothing$ imply $head(r)\cap S\neq\varnothing$. $S$ satisfies a ground program if $S$ satisfies every rule in $\Pi$. Let $\Pi$ be a program such that $\forall r\in\Pi,~body^-(r)=\varnothing$. Then, a set $S\subset Lit$ is a consistent answer set of $\Pi$---denoted $S\in AS(\Pi)$---if $S$ is a minimal set such that $S$ satisfies every rule from the ground instantiation of $\Pi$ and $S$ does not contain a pair of complementary literals. Next, let $\Pi$ be any extended disjunctive program and $S\subseteq Lit$. For every rule $r$ in the ground instantiation of $\Pi$, the rule $r^S:head(r)\leftarrow body^+(r)$ is included in the reduct $\Pi^S$ if $body^-(r)\cap S=\varnothing$. Then, $S\in AS(\Pi)$ if $S\in AS(\Pi^S)$.

    \begin{proposition}\label{pro:ASP}
        Given a program $\Pi$, a set $S\in AS(\Pi)$, and a literal $\rho$, $\rho\in S\implies \exists r\in\Pi$, $head(r)=\rho$, $body^{+}(r)\subseteq S$, and $body^{-}(r)\cap S=\varnothing$.
    \end{proposition}
    
    Actual causality being the focus of this paper, we will mainly detail the translation from the actual causality semantics defined in Section \ref{sec:Actual_cause} to the program $\pi_{\mathbb{C}}$ in ASP. $\pi_{\mathbb{C}}$ corresponds to the \textit{causal motor} in Figure \ref{image:ACE_structure}. With respect to the translation from the action language semantics introduced in Section \ref{sec:Modele} to the program $\pi_{\mathbb{A}}$ in ASP---corresponding to the \textit{event motor} in Figure \ref{image:ACE_structure}---we will only detail the rules that are essential for the understanding of $\pi_{\mathbb{C}}$\footnote{For interested readers, code is given in Appendix and \tt{https://gitlab.lip6.fr/sarmiento/jair2022}.}. Both $\pi_{\mathbb{A}}$ and $\pi_{\mathbb{C}}$ are independent of the domain---if the problem changes $\pi_{\mathbb{A}}$ and $\pi_{\mathbb{C}}$ do not need to be rewritten---and both use variables of three sorts: time variables of $\mathbb{T}$, fluent variables of $\mathbb{F}$, and event variables of $\mathbb{E}$. As shown in Figure \ref{image:ACE_structure}, two other programs are needed. $\pi_{sce}(\sigma)$ is the ASP program obtained by the translation of the scenario $\sigma\subseteq\mathbb{A}\times\mathbb{T}$. In this translation, each couple of timed actions $(a,t)\in\sigma$ is represented by a predicate $performs(a,t)$. $\pi_{con}(\kappa)$ is the ASP program obtained by the translation of the context $\kappa\stackrel{def}{=}(\mathbb{E},\mathbb{F},pre,tri,eff,S_0,>,\mathbb{T})$. As for $\pi_{\mathbb{A}}$, we will only detail the elements of this translation that are essential for the understanding of $\pi_{\mathbb{C}}$. The set $\mathbb{T}$ is represented by $time(0..N)$. Each fluent $f\in\mathbb{F}$ is represented by a predicate $fluent(f)$ and each fluent $f\in S_0$ is represented by a predicate $initially(f)$. Each action and exogenous event in $\mathbb{E}$ is respectively translated as $action(action\_name,action\_pre,action\_eff)$ and $auto(exog\_event\_name,exog\_event\_tri,exog\_event\_eff)$, where $action\_name\in\mathbb{A}$, $exog\_$ $event\_name\in\mathbb{U}$, $action\_pre$ a goal descriptor (GD) allowing to identify the preconditions of events as $pre$, $exog\_event\_tri$ a goal descriptor allowing to identify the triggering conditions of exogenous events as $tri$, and $action\_eff$ and $exog\_event\_eff$ goal descriptors allowing to identify the effects of events as $eff$. The predicates $conj(GD)$, $disj(GD)$, and $in(GD,GD\_L)$ allow to construct formulas of $\mathcal{P}$. The predicates $conj(GD)$ and $in(GD,GD\_L)$ allow to construct formulas of $\mathcal{E}$.
    
    \begin{example}[\textit{pollution---translation}]\label{ex:formalisation_ASP}
        Let's illustrate what an event translation looks like by taking the exogenous event discharging wastewater ($dis_w \in \mathbb{U}$) from Example \ref{ex:formalisation}. As a reminder, the event $dis_w$ is triggered either when there is wastewater from the speakers factory ($w_s$) or when there is wastewater from the medicine factory ($w_m$) and $w_m$ treatment plant is out of service ($t_{os}$). This event effect is to raise the pollution indicator of the lake above the threshold ($s_{sup}$). In our action language semantics we formalised it as $tri(dis_w)=w_s\vee(w_m\wedge t_{os}), eff(dis_w)=s_{sup}$. In $\pi_{con}(\kappa)$ it would be represented as:
        \begin{equation*}
        \begin{aligned}[c]
            &auto(dis_w,dis_wCond,dis_wEff).\\
            &disj(dis_wCond).\\
            &in(dis_wCond,w_s).\\
            &in(dis_wCond,dis_wCond1).\\
            &conj(dis_wCond1).
        \end{aligned}
        \hspace{2cm}
        \begin{aligned}[c]
            &in(dis_wCond1,w_m).\\
            &in(dis_wCond1,t_{os}).\\
            &conj(dis_wEff).\\
            &in(dis_wEff,s_{sup}).
        \end{aligned}
        \end{equation*}
    \end{example}
    
    \subsection*{The Program $\pi_{\mathbb{A}}$}
    
        The first group of rules determine the predicate $holds(F,T)$ indicating $f\in S^{\chi}(t)$ with $f\in\mathbb{F}$. \eqref{eq:holds_S0} states that $f\in S^{\chi}(0)$ if $\pi_{con}(\kappa)\vDash initially(f)$. \eqref{eq:holds_ini} indicates that $f\in S^{\chi}(t+1)$ if it has been initiated at time $t$. Finally, \eqref{eq:holds_ter} indicates that $f\in S^{\chi}(t+1)$ if $f\in S^{\chi}(t)$ and it has not been terminated by any event---this rule is motivated by the commonsense law of inertia.
    
        \begin{align}
            holds(F,0) :-~ &initially(F), fluent(F).\label{eq:holds_S0}\\
            holds(F,T+1) :-~ &initiated(E,F,T).\label{eq:holds_ini}\\
            holds(F,T+1) :-~ &holds(F,T), fluent(F), time(T),\nonumber\\
            &not~terminated(E,F,T):event(E).\label{eq:holds_ter}
        \end{align}
        
        In \citeauthor{wright_causation_1985}'s conception of causality \citep{wright_causation_1985}, causality can only be sufficient if we take into account---in addition to the positive causes---the conditions that were not true and whose absence was a necessary condition for the occurrence of the result---negative causes. Then, the events being causes of their absence are also causes of the result. By working on fluent literals, our definition of causation already takes this notion into account. The following rules extend the predicates $holds$ and $initially$ to manage fluent literals. \eqref{eq:initially_neg} states that $\pi_{con}(\kappa)\vDash initially(\neg f)$ if $\pi_{con}(\kappa)\not\vDash initially(f)$. \eqref{eq:holds_neg} indicates that $\neg f\in S^{\chi}(t)$ if $f\not\in S^{\chi}(t)$.
        
        \begin{align}
            initially(neg(F)) :-~ &not~initially(F), fluent(F).\label{eq:initially_neg}\\
            holds(neg(F),T) :-~ &not~holds(F,T), fluent(F), time(T).\label{eq:holds_neg}
        \end{align}
        
        The next group of rules takes over the changes that will take place at the level of fluents. \eqref{eq:initiated} states that the event $e\in E^{\chi}(t)$ has initiated the truthfulness of fluent $f$---$initiated(E,F,T)$---if the effects $Effect=eff(e)$ are applied, $f\in eff(e)$, and $f\not\in S^{\chi}(t)$. \eqref{eq:terminated} indicates that the event $e\in E^{\chi}(t)$ has terminated the truthfulness of fluent $F$---$terminated(E,F,T)$---if the effects $Effect=eff(e)$ are applied, $\neg f\in eff(e)$, and $f\in S^{\chi}(t)$. One of the roles of the predicate $apply(E,Effect,T)$ is to ensure $e\in E^{\chi}(t)$ which has many implications. For example, given that according to Definition \ref{def:valid} and \ref{def:traces}, $\tau_{\sigma,\kappa}^e$ is valid in $\kappa$, it implies $S^{\chi}(t)\vDash tri(e)$ and $\neg\exists e'\in E^{\chi}(t), e'>e$\footnote{The corresponding rules in ASP can be found in Appendix.}. This is ensured for both actions and exogenous events by the predicate $happens(E,GD,T)$ in rules \eqref{eq:apply_actions} and \eqref{eq:apply_exogenous}.
        
        \begin{align}
            initiated(E,F,T) :-~ &apply(E,Effect,T), in(Effect,F),\nonumber\\ 
            &not~holds(F,T), fluent(F).\label{eq:initiated}\\
            terminated(E,F,T) :-~ &apply(E,Effect,T), in(Effect,neg(F)),\nonumber\\
            &holds(F,T), fluent(F).\label{eq:terminated}
        \end{align}
        
        Together, predicates $initiated(E,F,T)$ and $terminated(E,F,T)$ occupy the equivalent role in $\pi_{\mathbb{A}}$ than $actualEff$ in our action language semantics since, given an $e\in E^{\chi}(t)$ and a state $S^{\chi}(t)$, the two conditions stated in Definition \ref{def:actualEff}---(i) $l_i\in eff(e)$ and (ii) $l_i\not\in S^{\chi}(t)$---are satisfied. Hence, $initiated(E,F,T)$ and $terminated(E,F,T)$ together are the translation into ASP of $actualEff(e,S^{\chi}(t))$ with $e\in E^{\chi}(t)$.
        
        \begin{align}
            apply(A,Effect,T) :- &happens(A,GD,T), action(A,GD,Effect).\label{eq:apply_actions}\\
            apply(U,Effect,T) :- &happens(U,GD,T), auto(U,GD,Effect).\label{eq:apply_exogenous}
        \end{align}
    
    \subsection*{The Program $\pi_{\mathbb{C}}$}
        
        In program $\pi_{\mathbb{C}}$, we represent the occurrence of events $(e,t)\in\mathbb{E}\times\mathbb{T}$ by the predicate $o(e,t)$ and the truthfulness of $\mathcal{P}$ formulas $(\psi,t)\in\mathbb{F}\times\mathbb{T}$ by the predicate $h(\psi,t)$. 
        
        \subsubsection*{Fluent to Fluent}
            
            The first group of rules determine basic relations between elements of $\mathbb{F}$. \eqref{eq:inertia} states that there is a relation of inertia between $h(l,t)$ and $h(l,t+1)$ if $l\in Lit_\mathbb{F}$, $l\in S^{\chi}(t)$, and $l\in S^{\chi}(t+1)$. This rule is necessary given our adherence to common sense law of inertia. 
            
            \begin{equation}
                inertia(h(L,T),h(L,T+1)) :-~ holds(L,T), holds(L,T+1), literal(L).\label{eq:inertia}
            \end{equation}
            
            \eqref{eq:r_hh_conj} indicates that there is a relation between $h(\psi,t)$ and $h(\psi_\wedge,t)$ if $\psi_\wedge$ is a conjunction---hence the notation---$\psi_\wedge\in S^{\chi}(t)$, and $\psi\in\psi_\wedge$. Notice the use of the variable $GD\_L$ as $\psi$ given the method by which formulas are constructed in $\pi_{con}(\kappa)$. \eqref{eq:r_hh_disj} states that there is a relation between $h(\psi,t)$ and $h(\psi_\vee,t)$ if $\psi_\vee$ is a disjunction---hence the notation---$\psi_\vee\in S^{\chi}(t)$, $\psi\in\psi_\vee$, and $\psi\in S^{\chi}(t)$.
            
            \begin{align}
                r\_hh(h(GD\_L,T),h(C,T)) :-~ &conj(C), holds(C,T), in(C,GD\_L).\label{eq:r_hh_conj}\\
                r\_hh(h(GD\_L,T),h(D,T)) :-~ &disj(D), holds(D,T), in(D,GD\_L), holds(GD\_L,T).\label{eq:r_hh_disj}
            \end{align}
        
        \subsubsection*{Direct NESS-cause}
            
            The next group of rules determine the direct NESS-cause relations as defined in Definition \ref{def:direct_NESS}. \eqref{eq:direct_S0} states that $o(ini_l,-1)$ is a direct NESS-cause of $h(l,0)$ if $l\in S^{\chi}(0)$. In this way, we are faithful to the philosophical idea that a causal chain can be backtracked indefinitely by making the events which are beyond our bounded past formalisation appear in the causal relations.
            
            \begin{equation}
                direct\_ness(o(ini(L),-1),h(L,0)) :-~ initially(L).\label{eq:direct_S0}
            \end{equation}
            
            \eqref{eq:direct_ini} indicates that $o(e,t)$ is a direct NESS-cause of $h(f,t+1)$ if $f\in actualEff(e,S^{\chi}(t))$ with $e\in E^{\chi}(t)$. Similarly, \eqref{eq:direct_ter} states that $o(e,t)$ is a direct NESS-cause of $h(\neg f,t+1)$ if $\neg f\in actualEff(e,S^{\chi}(t))$ with $e\in E^{\chi}(t)$. These rules formalise the idea mentioned in Section \ref{sec:Modele} according to which the predicate $actualEff$ can be equated to basic causal information given by our action language semantics. 
            
            \begin{align}
                direct\_ness(o(E,T),h(F,T+1)) :-~ &initiated(E,F,T).\label{eq:direct_ini}\\
                direct\_ness(o(E,T),h(neg(F),T+1)) :-~ &terminated(E,F,T).\label{eq:direct_ter}
            \end{align}
            
            \eqref{eq:direct_inertia} indicates that the occurrence of event denoted $Event$ is a direct NESS-cause of $h(l,t+1)$ if $Event$ is a direct NESS-cause of $h(l,t)$ and there is a relation of inertia between $h(l,t)$ and $h(l,t+1)$. This rule reflects transitivity by inertia: if $(e,t)$ is a direct NESS-cause of $(l,t)$, it will be a direct NESS-cause of the truthfulness of $l$ at all time points after $t$ until $l$ is made false. The use of the $Event$ variable makes the transitivity relation more clear since it emphasises that we do not need to define a clear relation between the time of the occurrence of event and the truthfulness of $l$.
            
            \begin{align}
                direct\_ness(Event,h(L,T+1)) :-~ &direct\_ness(Event,h(L,T)),\nonumber\\
                &inertia(h(L,T),h(L,T+1)).\label{eq:direct_inertia}
            \end{align}
            
            Finally, \eqref{eq:direct_formula} states that the occurrence of event denoted $Event$ is a direct NESS-cause of $h(\psi,t)$ if $Event$ is a direct NESS-cause of $h(\psi',t)$ and there is a relation between $h(\psi',t)$ and $h(\psi,t)$. This rule also reflects transitivity, but this time with respect to logical connectives and in a same time point.
            
            \begin{align}
                direct\_ness(Event,h(GD,T)) :-~ &direct\_ness(Event,h(GD\_L,T)),\nonumber\\
                &r\_hh(h(GD\_L,T),h(GD,T)).\label{eq:direct_formula}
            \end{align}
            
            Notice that although there is transitivity by inertia for literals, there is no transitivity by inertia for more complex formulas. Using Example \ref{ex:switches} in a slightly different way shows that this would not be desirable. Indeed, making $ini_{l_4}\in E^{\chi}(-1)$ would make true $\psi$ at all time points. If transitivity by inertia was allowed for formulas such as $\psi$, our program will consider $(ini_{l_1},-1)$ as a direct NESS-cause of $(\psi,2)$ which we discussed was factually not correct, and thus not in accordance with our definition.
            
            \begin{definition}[actual causality program $\Pi(\chi)$]
                Given a scenario $\pi_{sce}(\sigma)$, a context $\pi_{con}(\kappa)$, an event motor $\pi_{\mathbb{A}}$, and a causal motor $\pi_{\mathbb{C}}$, the actual causality program is $\Pi(\chi) = \pi_{sce}(\sigma)\cup\pi_{con}(\kappa)\cup\pi_{\mathbb{A}}\cup\pi_{\mathbb{C}}$.
            \end{definition}
    
            \begin{theorem}[direct NESS completeness and soundness]\label{the:complete_and_sound}
                Given a causal setting $\chi$ and a DNF, subsumption minimal\footnote{$\psi_{DNF}=\psi_1 \vee \dots \vee \psi_m$ being the disjunctive normal form of $\psi$, $\forall i,j\in\left\{1,\dots,m\right\}^2,i\neq j, \psi_i\not\vDash\psi_j$.}, and tautology free formula $\psi\in\mathcal{P}$ such that $\psi\not\vDash\bot$, $(e,t)\in\mathbb{E}\times\mathbb{T}$ is a direct NESS-cause of $(\psi,t_\psi)$, iff $\Pi(\chi)\vDash direct\_ness(o(e,t),h(\psi,t_\psi))$\footnote{Notice that the last $\vDash$ corresponds to entailment in the sense of logic programs which needs to be distinguished from the one introduced in Section \ref{sec:Modele}.}.
            \end{theorem}

        \subsubsection*{NESS-cause}
            
            The next group of rules determine the NESS-cause relations as defined in Definition \ref{def:NESS}. \eqref{eq:direct_is_ness} states that $o(e_1,t_1)$ is a NESS-cause of $h(\psi,t_\psi)$ if $o(e_1,t_1)$ is a direct NESS-cause of $h(\psi,t_\psi)$---direct NESS-causes are NESS-causes. This corresponds to the base case of Definition \ref{def:NESS}.
            
            \begin{align}
                ness(o(E1,T1),h(GD\_L,T2)) :-~direct\_ness(o(E1,T1),h(GD\_L,T2)).\label{eq:direct_is_ness}
            \end{align}
            
            \eqref{eq:ness_trig} indicates that $o(e_1,t_1)$ is a NESS-cause of $h(\psi,t_\psi)$ if $o(e_1,t_1)$ is an actual cause of $o(e_2,t_2)$ and $o(e_2,t_2)$ is a NESS-cause of $h(\psi,t_\psi)$. This corresponds to the recursive case of Definition \ref{def:NESS}. Those are the cases where we ask ourselves: why the identified causes---corresponding to $o(e_2,t_2)$---had been triggered? More precisely, it corresponds to the case where $o(e_1,t_1)$ is a NESS-cause of $tri(C_R(t_i))$ in Definition \ref{def:NESS}.  
            
            \begin{align}
                ness(o(E1,T1),h(GD\_L,T3)) :-~ &actual(o(E1,T1),o(E2,T2)),\nonumber\\
                &ness(o(E2,T2),h(GD\_L,T3)).\label{eq:ness_trig}
            \end{align}
        
            \begin{theorem}[NESS completeness and soundness]\label{the:complete_and_sound_1}
                Given a causal setting $\chi$ and a DNF, subsumption minimal, and tautology free formula $\psi\in\mathcal{P}$ such that $\psi\not\vDash\bot$, $(e,t)\in\mathbb{E}\times\mathbb{T}$ is a NESS-cause of $(\psi,t_\psi)$, iff $\Pi(\chi)\vDash ness(o(e,t),h(\psi,t_\psi))$.
            \end{theorem}
        
        \subsubsection*{Actual Cause}
        
            The next group of rules determine the actual cause relation as defined in Definition \ref{def:actual}. \eqref{eq:actual_ness} states that $o(e_1,t_1)$ is an actual cause of $o(e_2,t_2)$ if $o(e_1,t_1)$ is a NESS-cause of $h(\psi,t_2)$, $tri(e_2)=\psi$, and $e_2\in\mathbb{U}$.
            
            \begin{align}
                actual(o(E1,T1),o(E2,T2)) :-~ &ness(o(E1,T1),h(GD,T2)),\nonumber\\
                &happens(E2,GD,T2), auto(E2,GD,Eff).\label{eq:actual_ness}
            \end{align}
    
            \begin{theorem}[actual cause completeness and soundness]\label{the:complete_and_sound_2}
                Given a causal setting $\chi$, a DNF, subsumption minimal, and tautology free formula $\psi\in\mathcal{P}$ such that $\psi\not\vDash\bot$ and $tri(e')=\psi$, $(e,t)\in\mathbb{E}\times\mathbb{T}$ is an actual cause of $(e',t')\in\mathbb{E}\times\mathbb{T}$, iff $\Pi(\chi)\vDash actual(o(e,t),o(e',t'))$.
            \end{theorem}

    \section{Conclusion}
\label{sec:conclude}
    
    In this paper we present a dual contribution. First, we strengthen the link built in \citep{sarmiento_action_2022-1} between automated planning and causality by proposing a complete and sound translation into logic programming from an actual causation definition suitable for action languages. Second, we link computational ethics and causality by showing the necessity of a mechanism allowing to establish complex causal relations in the simulation of ethical reasoning---a practice that is still rare in the domain---and by proposing such a mechanism. Based on the actual causation definition we established, we ensured that our resulting program addresses two of what we consider the main remaining limitations of the linking automated planning and causality venture. First, not to yield into the easy confusion between responsibility and causality, our proposal is suitable for a factual and independent of policy choices causal inquiry. Second, not to disregard the much debated cases of overdetermination, our proposal is based on an action language semantics allowing event concurrency. By taking as a basis Wright's NESS test we are able to satisfactorily manage these cases. To the best of our knowledge, no other actual causality logic programming has been able to handle those complex cases, albeit essential. Our approach thus enables handling complex cases of causality in decision-making applications. Integrating this approach into an ethical decision-making framework allows us by the same occasion to contribute to the progress of computational ethics by enabling us to deal with situations that were previously out of reach.
    
    In future work we intend to extend the definition of causality by including the relation `prevent' \citep{berreby_event-based_2018}. This relation refers to the case where an event did not occur because a condition that was not true and whose absence was a necessary condition for the occurrence of the result was made true. From that perspective, the events being causes of such a condition are causes of the non-occurrence of the result. As approaches based on structural equations do not distinguish events from fluents, they cannot be used to distinguish causing the negation of a fluent from causing an event not to occur---a far more complex relation. This relation seems to be indispensable in ethical reasoning. In the example used throughout this paper, if an agent had the possibility to prevent the harm, not having averted it is a relevant information for the ethical evaluation. This suggests that we should continue down the path of action languages by defining the `prevent' relationship given our more complex framework allowing for events concurrency and disjunction.
    
    Furthermore, considering the implications for ethical evaluation, it appears essential to investigate in future work the difference between the concepts of `enables' and `cause'---a point on which some interesting leads from various perspectives have been given \citep{martin_dispositions_1994,lewis_finkish_1997,sloman_causal_2009,berreby_event-based_2018,choi_dispositions_2021}. Consider that the precondition for the action burglarising a house is that the door is left open: if an agent decides to go and rob the house, it seems correct to say that another agent who forgot to close the door has enabled the robbery, not that he is an actual cause of it. Indeed, in this case the event of interest is an action of an agent with a volition. The `enables' relation seems to be definable by replacing the $tri$ function by $pre$ in the actual cause definition. The possible nuances of causality in relation with the `enabling' notion do not seem to have been investigated by analytical philosophy in depth---considering all possible theoretical implications. However, applied bioethics seems to be an exciting theoretical ground in which the difference between causing and enabling an event appears crucial and even well conceptualised. This is particularly true when we deal with end-of-life issues such as euthanasia and assisted suicide. As numerous bioethics committees have shown, the crucial distinction between killing and letting die is based upon the assumption of different kind of moral responsibility when causing or allowing an event to happen. Thus, we intend to start from these issues to open a path of research that could definitively enrich our causality theory.

    \bibliographystyle{plainnat}
\bibliography{main}

    \newpage
    
    \appendix
\label{sec:Annexes}
\setcounter{proposition}{0}
\setcounter{theorem}{0}

\section*{Proofs of Direct NESS-causes Propositions}

    \begin{proposition}[direct NESS-cause of a literal]
        Given a causal setting $\chi$, $e\in E^{\chi}(t)$, and a fluent literal $l\in Lit_{\mathbb{F}}$ true at $t_l$, $(e,t)$ is a direct NESS-cause of $(l,t_l)$ iff ($c_1$) $S^{\chi}(t)\triangleright \left\{e\right\}\vDash l$, ($c_2$) $\forall t',~t<t'\leq t_l$, $S^{\chi}(t')\vDash l$, and ($c_3$) $S^{\chi}(t)\not\vDash l$.
    \end{proposition}

    \begin{proof}
        $[\implies]$
            Let $(e,t)$ be a direct NESS-cause of $(l,t_l)$. By Definition \ref{def:direct_NESS}, $\exists C\subseteq\mathbb{E}\times\mathbb{T}$, such that $C\underset{W}{\rightsquigarrow}(l,t_l)$ and $(e,t)\in C$. The backing $W$ is necessarily the partial state $L=\left\{l\right\}$ since $L$ is the unique set satisfying the causal sufficiency and minimality of the backing condition for the causal relation, given that $L\vDash l$ and that its only subset---namely $\varnothing$---does not satisfy $l$.
            
            A partition $W_1\dots,W_k$ of $W$ satisfies by definition the conditions (i) $\forall i \in\left\{1,\dots,k\right\}, W_i\neq\varnothing$, (ii) $\bigcup_{i\in\left\{1,\dots,k\right\}}W_i=W$, and (iii) $\forall i,j\in\left\{1,\dots,k\right\}^2, i\neq j \implies W_i\cap W_j=\varnothing$. The singleton $W$ has exactly one partition $W_1=\left\{l\right\}$ to which we associate the time $t_1$, thus by Definition \ref{def:direct_NESS}, $C=C(t_1)$. Hence, since $(e,t)\in C$, $t_1=t$.  
            
            As the unique partition is $W_1=\left\{l\right\}$, persistency of necessity can be expressed in this case as $\forall t', t<t'\leq t_{l}, ~S^{\chi}(t') \vDash l$---that is ($c_2$). By the same reasoning, weak necessity and minimality of $C$ can be expressed as $S^{\chi}(t)\triangleright C(t)\vDash l$ and $\forall C'\subset C(t), ~S^{\chi}(t)\triangleright C'\not\vDash l$. The condition $S^{\chi}(t)\not\vDash l$---that is ($c_3$)---is included in the minimality condition since for the case where $C'= \varnothing$, the minimality condition is $S^{\chi}(t) \triangleright \varnothing \not\vDash l$ with $S^{\chi}(t) = S^{\chi}(t) \triangleright \varnothing$ by Definition \ref{def:update_op}. As $(e,t)\in C=C(t)$, $C(t)=\left\{(e,t)\right\} \cup C''$, where $C''=\left\{(e'',t), e''\in E^{\chi}(t)\right\}$. Hence, weak necessity can be written as $S^{\chi}(t)\triangleright(\left\{(e,t)\right\} \cup C'')\vDash l$, which by Definition \ref{def:actualEff} and \ref{def:update_op} means $l\in S^{\chi}(t)\setminus \left[\overline{actualEff(\left\{(e,t)\right\},S^{\chi}(t))\cup actualEff(C'',S^{\chi}(t))}\right]\cup actualEff(\left\{(e,t)\right\},S^{\chi}(t))\cup actualEff(C'',S^{\chi}(t))$. Three cases can be considered. Either $l$ belongs to:\\
            (i) $S^{\chi}(t)\setminus\overline{actualEff(\left\{(e,t)\right\},S^{\chi}(t))\cup actualEff(C'',S^{\chi}(t))}$;\\
            (ii) $actualEff(C'',S^{\chi}(t))$;\\
            or (iii) $actualEff(\left\{(e,t)\right\},S^{\chi}(t))$.\\
            The first case implies $l\in S^{\chi}(t)$, which is contradictory to the condition $S^{\chi}(t)\not\vDash l$. The second case implies $S^{\chi}(t)\triangleright C''\vDash l$, which is contradictory to the minimality condition $\forall C'\subset C(t), ~S^{\chi}(t)\triangleright C'\not\vDash l$ given that $C''\subset C(t)$. The third case is then the only possible case, it implies $l\in actualEff(\left\{(e,t)\right\},S^{\chi}(t))$, which can be written as $S^{\chi}(t)\triangleright \left\{e\right\}\vDash l$---that is ($c_1$). We have therefore proven that when $(e,t)$ is a direct NESS-cause of $(\psi,t_{\psi})$ we have ($c_1$), ($c_2$), and ($c_3$).
            
        $[\impliedby]$ 
            Let $l$ be a fluent literal satisfied at $t_l$, $(e,t)$ be an event occurrence satisfying ($c_1$), ($c_2$), and ($c_3$), and $C=\left\{(e,t)\right\}$. Consider an hypothetical backing $W=\left\{l\right\}$ of the truthfulness of $l$ at $t_l$. This backing $W$ satisfies the causally sufficient and minimality conditions since $\left\{l\right\}\vDash l$ and that its only subset---namely $\varnothing$---does not satisfy $l$.
            
            There is a decreasing sequence $t_1=t$ and a partition $W_1=\left\{l\right\}$ of $W$ such that $\forall i\in\left\{1\right\}$, given $C(t_i) = C \cap E^{\chi}(t_i)$: as $C=\left\{(e,t_i)\right\}$, $C=\bigcup_{i\in\left\{1,\dots,k\right\}} C(t_i)$, which corresponds to minimality of $C$, ($c_1$) can be rewritten as $S^{\chi}(t_i)\triangleright C(t_i)\vDash W_i$, which corresponds to weak necessity of $C$, ($c_2$) can be rewritten as $\forall t', t_i<t'\leq t_{l},~S^{\chi}(t') \vDash W_i$, which corresponds to persistency of necessity, and ($c_3$) can be rewritten as $S^{\chi}(t_i)\not\vDash W_i$, which corresponds to minimality of $C$ at $t_i$ since it implies $S^{\chi}(t_i)\triangleright\varnothing\not\vDash W_i$ with $\varnothing$ being the unique proper subset of $\left\{e\right\}$. Therefore, $\exists C\subseteq\mathbb{E}\times\mathbb{T}$ such that $(e,t)\in C$ and $C\rightsquigarrow(l,t_{l})$.
    \end{proof}

    \begin{proposition}[direct NESS-cause of a literal conjunction]
        Given a causal setting $\chi$, $e\in E^{\chi}(t)$, and a formula $\psi = l_1 \wedge \dots \wedge l_m$ of $\mathcal{P}$ true at $t_\psi$, $(e,t)$ is a direct NESS-cause of $(\psi,t_{\psi})$ iff ($c_1$) $\exists j\in \left\{1,\dots,m\right\},~S^{\chi}(t)\triangleright \left\{e\right\}\vDash l_j$, ($c_2$) $\forall t',~t<t'\leq t_\psi$, $S^{\chi}(t')\vDash l_j$, and ($c_3$) $S^{\chi}(t)\not\vDash l_j$.
    \end{proposition}

    \begin{proof}
        $[\implies]$
            Let $(e,t)$ be a direct NESS-cause of $(\psi,t_{\psi})$. By Definition \ref{def:direct_NESS}, $\exists C\subseteq\mathbb{E}\times\mathbb{T}$, such that $C\underset{W}{\rightsquigarrow}(\psi,t_{\psi})$ and $(e,t)\in C$. The backing $W$ is necessarily the partial state $L=\left\{l_1, \dots, l_m\right\}$ since $L$ is the unique set satisfying the causal sufficiency and minimality of the backing condition for the causal relation, given that $L\vDash\psi$ and that $\forall L',~L'\subset L, L'\not\vDash\psi$ since $\forall L',~\exists l_j\in\psi,~l_j\not\in L'$.
            
            For each $l_j$, let $t_{l_j}$ be the unique time in which $S^{\chi}(t_{l_j})\not\vDash l_j$ and $\forall t',~t_{l_j}<t'\leq t_\psi$, $S^{\chi}(t')\vDash l_j$, time which necessarily exists since we now that $S^{\chi}(t_\psi)\vDash \psi \implies S^{\chi}(t_\psi)\vDash l_j$ and that $S^{\chi}(-1)\not\vDash l_j$ given that $S^{\chi}(-1)=\varnothing$. Consider $T=\left\{t_{l_j},j\in\left\{1,\dots,m\right\}\right\}$ and $t_1,\dots,t_k$ the decreasing sequence corresponding to $T$, where $k\leq m$. The partition of the backing $W$ can then be defined as $\forall i\in\left\{1,\dots,k\right\}, W_i=\left\{l_j\in W, t_{l_j}=t_i\right\}$.
            
            As $(e,t)\in C$ and---by Definition \ref{def:direct_NESS}---$C(t_i) = C \cap E^{\chi}(t_i)$, $\exists i\in\left\{1,\dots,k\right\}$, $t_i=t$ and $S^{\chi}(t)\triangleright C(t)\vDash W_i$ with $C(t)=\left\{(e,t)\right\} \cup C''$, where $C''=\left\{(e'',t), e''\in E^{\chi}(t)\right\}$. Hence, weak necessity of $C$ can be rewritten as $\forall l_j\in W_i,~S^{\chi}(t)\triangleright(\left\{(e,t)\right\} \cup C'')\vDash l_j$, which by Definition \ref{def:update_op} means $\forall l_j\in W_i,~l_j\in S^{\chi}(t)\setminus\left[\overline{actualEff(\left\{(e,t)\right\},S^{\chi}(t))\cup actualEff(C'',S^{\chi}(t))}\right]\cup actualEff(\left\{(e,t)\right\},S^{\chi}(t))\cup actualEff(C'',S^{\chi}(t))$. Three cases can be considered. Either $l_j$ belongs to :\\
            (i) $S^{\chi}(t)\setminus\overline{actualEff(\left\{(e,t)\right\},S^{\chi}(t))\cup actualEff(C'',S^{\chi}(t))}$;\\
            (ii) $actualEff(\left\{(e,t)\right\},S^{\chi}(t))$;\\
            or (iii) $actualEff(C'',S^{\chi}(t))$.\\
            The first case implies $l_j\in S^{\chi}(t)$, which is contradictory to the condition $S^{\chi}(t)\not\vDash l_j$---that is ($c_3$)---inherent by the construction of the partition of the backing, thus $l_j$ necessarily belongs to $actualEff(\left\{(e,t)\right\},S^{\chi}(t)) \cup actualEff(C'',S^{\chi}(t))$. The case where $\forall l_j\in W_i, l_j\not\in actualEff(\left\{(e,t)\right\},S^{\chi}(t))$ is contradictory to the minimality condition $\forall C'\subset C(t), ~S^{\chi}(t)\triangleright C'\not\vDash W_i$ given that it implies $\forall l_j\in W_i, l_j\in actualEff(C'',S^{\chi}(t))$ and that by Definition \ref{def:direct_NESS} we have $\left\{(e,t)\right\}\subset C(t)$. The case where $\exists j\in \left\{1,\dots,m\right\},~l_j\in actualEff(\left\{(e,t)\right\},S^{\chi}(t))$ is then the only possible case. By Definition \ref{def:update_op}, this result can be rewritten as $\exists j\in \left\{1,\dots,m\right\},~S^{\chi}(t)\triangleright \left\{e\right\}\vDash l_j$---that is ($c_1$). Finally, for this same $i\in\left\{1,\dots,k\right\}$ such that $t_i=t$, the persistency of necessity condition can be rewritten as $\forall l_j\in W_i, \forall t', ~t<t'\leq t_\psi, ~S^{\chi}(t')\vDash l_j$---that is ($c_2$). We have therefore proven that when $(e,t)$ is a direct NESS-cause of $(\psi,t_{\psi})$ we have ($c_1$), ($c_2$), and ($c_3$).

        $[\impliedby]$
            Let $\psi$ be a literal conjunction satisfied at $t_\psi$ and $(e,t)$ be an event occurrence satisfying ($c_1$), ($c_2$), and ($c_3$). Consider an hypothetical backing $W=\left\{l_1, \dots, l_m\right\}$ of the truthfulness of $\psi$ at $t_\psi$. This backing $W$ satisfies the causally sufficient and minimality conditions since $\left\{l_1, \dots, l_m\right\}\vDash\psi$ and that $\forall W',~W'\subset W, W'\not\vDash\psi$ since $\forall W',~\exists l_j\in\psi,~l_j\not\in W'$.
            
            As $S^{\chi}(t_\psi)\vDash \psi$ and $S^{\chi}(-1)=\varnothing$, $\exists C\subseteq\mathbb{E}\times\mathbb{T}$ such that $C\underset{W}{\rightsquigarrow}(\psi,t_\psi)$. Given that ($c_1$) $\exists j\in \left\{1,\dots,m\right\},~S^{\chi}(t)\triangleright \left\{e\right\}\vDash l_j$, ($c_2$) $\forall t',~t<t'\leq t_\psi$, $S^{\chi}(t')\vDash l_j$, and ($c_3$) $S^{\chi}(t)\not\vDash l_j$---according to Proposition \ref{pro:direct_NESS_literal}---$(e,t)$ is a direct NESS-cause of $(l_j,t_{\psi})$. Hence, by Definition \ref{def:direct_NESS}, $\exists C'\subseteq\mathbb{E}\times\mathbb{T}$ such that $(e,t)\in C'$ and $C'\underset{W_j}{\rightsquigarrow}(l_j,t_\psi)$. As shown in proof of Proposition \ref{pro:direct_NESS_literal}, the backing of this causal relation is necessarily $W_j=\left\{l_j\right\}$. Since $l_j\in\psi$ and the partition of $W$ satisfies by definition $\bigcup_{i\in\left\{1,\dots,k\right\}}W_i=W$, $\exists i,W_j\subseteq W_i$ and $t_i=t$. Therefore, $C'\subseteq C(t)$ and thus $(e,t)\in C$.
    \end{proof}

    \begin{proposition}[direct NESS-cause of a DNF]
        Given a causal setting $\chi$, $e\in E^{\chi}(t)$, and a DNF, subsumption minimal, and tautology free formula $\psi = \psi_1 \vee \dots \vee \psi_m$ of $\mathcal{P}$ such that $\psi\not\vDash\bot$, $(e,t)$ is a direct NESS-cause of $(\psi,t_{\psi})$ iff ($c_1$) $\exists j\in \left\{1,\dots,m\right\}$, $S^{\chi}(t_\psi)\vDash \psi_j$, ($c_2$) $S^{\chi}(t)\triangleright \left\{e\right\}\vDash l\in\psi_j$, ($c_3$) $\forall t',~t<t'\leq t_\psi$, $S^{\chi}(t')\vDash l$, and ($c_4$) $S^{\chi}(t)\not\vDash l$.
    \end{proposition}

    \begin{proof}
        $[\implies]$
            Let $(e,t)$ be a direct NESS-cause of $(\psi,t_{\psi})$. By Definition \ref{def:direct_NESS}, $\exists C\subseteq\mathbb{E}\times\mathbb{T}$, such that $C\underset{W}{\rightsquigarrow}(\psi,t_{\psi})$ and $(e,t)\in C$. Given $\psi$ a DNF, subsumption minimal, and tautology free formula, the possible backings $W_1,\dots,W_m$ are necessarily the partial states $L_1,\dots,L_m$ corresponding to each of the literal conjunctions $\psi_1,\dots,\psi_m$---as in the proof of Proposition \ref{pro:direct_NESS_conjunction}---since those partial states are all possible causally sufficient and minimal backings of the causal relation. The causal relation $C\underset{W}{\rightsquigarrow}(\psi,t_{\psi})$ is concerned with only one of those backings. Hence, $\exists j\in \left\{1,\dots,m\right\}, W_j=W$ and $S^{\chi}(t_\psi)\vDash \psi_j$---that is ($c_1$)---with $\psi_j = l_1 \wedge \dots \wedge l_n$ being the disjunct associated to $W_j$. The existing causal relation can then be rewritten as $C\underset{W_j}{\rightsquigarrow}(\psi_j,t_{\psi})$. According to Proposition \ref{pro:direct_NESS_conjunction}, given that $(e,t)\in C$ and $\psi_j$ a literal conjunction, $\exists i\in \left\{1,\dots,n\right\},~S^{\chi}(t)\triangleright \left\{e\right\}\vDash l_i\in\psi_j$---that is ($c_2$)---$\forall t',~t<t'\leq t_\psi$, $S^{\chi}(t')\vDash l_i$---that is ($c_3$)---and $S^{\chi}(t)\not\vDash l_i$---that is ($c_4$). We have therefore proven that when $(e,t)$ is a direct NESS-cause of $(\psi,t_{\psi})$ we have ($c_1$), ($c_2$), ($c_3$), and ($c_4$).
            
        $[\impliedby]$
            Let $\psi$ be a DNF satisfied at $t_\psi$ and $(e,t)$ be an event occurrence satisfying ($c_1$), ($c_2$), ($c_3$), and ($c_4$). Consider hypothetical backings $W_1,\dots,W_m$ of the truthfulness of $\psi$ at $t_\psi$, each corresponding to each of the literal conjunctions $\psi_1,\dots,\psi_m$. According to the proof of Proposition \ref{pro:direct_NESS_conjunction} and that $\psi$ a DNF, subsumption minimal, and tautology free formula, the backings $W_1,\dots,W_m$ all satisfy the causally sufficient and minimality conditions.
            
            As ($c_1$) $\exists j\in \left\{1,\dots,m\right\},~S^{\chi}(t_\psi)\vDash \psi_j$ and $S^{\chi}(-1)=\varnothing$, $\exists C\subseteq\mathbb{E}\times\mathbb{T}$ such that $C\underset{W_j}{\rightsquigarrow}(\psi_j,t_{\psi})$. Given that ($c_2$) $S^{\chi}(t)\triangleright \left\{e\right\}\vDash l\in\psi_j$, ($c_3$) $\forall t',~t<t'\leq t_\psi$, $S^{\chi}(t')\vDash l$, and ($c_4$) $S^{\chi}(t)\not\vDash l$---according to Proposition \ref{pro:direct_NESS_conjunction}---$(e,t)$ is a direct NESS-cause of $(\psi_j,t_{\psi})$. Hence, by Definition \ref{def:direct_NESS}, $(e,t)\in C$. Since $W_j$ satisfy the causally sufficient and minimality conditions when interested in the truthfulness of $\psi$ at $t_\psi$, the above causal relation implies $C\underset{W_j}{\rightsquigarrow}(\psi,t_\psi)$.
    \end{proof}

\section*{Proofs of Completeness and Soundness of Direct NESS-Causes}
\stepcounter{proposition}
\stepcounter{proposition}

    \begin{proposition}[Completeness if $\psi$ a literal]\label{pro:complete_literal}
        Given $(e,t)\in\mathbb{E}\times\mathbb{T}$ a direct NESS-cause of $(l,t_l)$, $\Pi(\chi)\vDash direct\_ness(o(e,t),h(l,t_l))$.
    \end{proposition}
    
    \begin{proof}
        Given $(e,t)$ a direct NESS-cause of $(l,t_l)$---according to Proposition \ref{pro:direct_NESS_literal}---($c_1$) $S^{\chi}(t)\triangleright \left\{e\right\}\vDash l$, ($c_2$) $\forall t',~t<t'\leq t_l$, $S^{\chi}(t')\vDash l$, and ($c_3$) $S^{\chi}(t)\not\vDash l$. Three cases can be considered.
        
        Case 1: In the particular case where $t_l=0$, the only way by which $S^{\chi}(0)\vDash l$ is that $l\in S_0$. Thus by construction of $\Pi(\chi)$, $\Pi(\chi)\vDash initially(l)$. Thus, given the rule:
        $$\eqref{eq:direct_S0}: direct\_ness(o(ini(L),-1),h(L,0))~:-~initially(L).$$
        we have $\Pi(\chi)\vDash direct\_ness(o(ini(l),-1),h(l,0))$ where $ini(l)=e$ and $t=-1$.
        
        Case 2: In the case where $t_l=t+1$, given $e\in E^{\chi}(t)$, ($c_1$), and ($c_3$), $\Pi(\chi)\vDash initiated(e,l,t)$ or $\Pi(\chi)\vDash terminated(e,\overline{l},t)$. Then, given rules:
        $$\eqref{eq:direct_ini}: direct\_ness(o(E,T),h(L,T+1))~:-~initiated(E,L,T).$$
        $$\eqref{eq:direct_ter}: direct\_ness(o(E,T),h(neg(L),T+1))~:-~terminated(E,L,T).$$
        we have $\Pi(\chi)\vDash direct\_ness(o(e,t),h(l,t+1))$.
        
        Case 3: Otherwise, we have $t_l>t+1$---follows from $t_l\neq 0$ and $t_l\neq t+1$ since $t_l>t$ because of our assumption that causes precedes effects in time. As before, we have $\Pi(\chi)\vDash direct\_ness(o(e,t),h(l,t+1))$. Additionally, given the condition $\forall t', ~t<t'\leq t_l, ~S^{\chi}(t')\vDash l$, $\Pi(\chi)\vDash holds(l,t')$, where $t<t'\leq t_l$. Then, given rule:
        $$\eqref{eq:inertia}: inertia(h(L,T),h(L,T+1)) :-~ holds(L,T), holds(L,T+1), literal(L).$$
        we have the complete chain from $\Pi(\chi)\vDash inertia(h(l,t+1),h(l,t+2))$ until $\Pi(\chi)\vDash inertia(h(l,t_l-1),h(l,t_l))$. Thus, given $\Pi(\chi)\vDash direct\_ness(o(e,t),h(l,t+1))$ and the rule:
            \begin{align*}
                \eqref{eq:direct_inertia}: direct\_ness(Event,h(L,T+1))~:-~&direct\_ness(Event,h(L,T)),\\ &inertia(h(L,T),h(L,T+1)).
            \end{align*}
        we have the complete chain from $\Pi(\chi)\vDash direct\_ness(o(e,t),h(l,t+2))$ until the desired result $\Pi(\chi)\vDash direct\_ness(o(e,t),h(l,t_l))$.
    \end{proof}

    \begin{proposition}[Completeness if $\psi$ a literal conjunction]\label{pro:complete_conjunction}
        Given $(e,t)\in\mathbb{E}\times\mathbb{T}$ a direct NESS-cause of $(\psi,t_\psi)$, where $\psi = l_1 \wedge \dots \wedge l_m$, $\Pi(\chi)\vDash direct\_ness(o(e,t),h(\psi,t_\psi))$.
    \end{proposition}
    
    \begin{proof}
        Given $(e,t)$ a direct NESS-cause of $(\psi,t_\psi)$---according to Proposition \ref{pro:direct_NESS_conjunction}---($c_1$) $\exists j\in \left\{1,\dots,m\right\},~S^{\chi}(t)\triangleright \left\{e\right\}\vDash l_j$, ($c_2$) $\forall t',~t<t'\leq t_\psi$, $S^{\chi}(t')\vDash l_j$, and ($c_3$) $S^{\chi}(t)\not\vDash l_j$.
        
        Considering the conjunction of literals $\psi = l_1 \wedge \dots \wedge l_m$ true at $t_\psi$, $\Pi(\chi)\vDash conj(\psi)$, $\Pi(\chi)\vDash in(\psi,l_1)$, $\dots$, $\Pi(\chi)\vDash in(\psi,l_m)$, and $\Pi(\chi)\vDash holds(\psi,t_\psi)$. Thus, given the rule:
        $$\eqref{eq:r_hh_conj}: r\_hh(h(L,T),h(C,T))~:-~conj(C), holds(C,T), in(C,L).\footnote{For the sake of clarity, since we now $\psi$ a conjunction of literals we replaced the variable $GD\_L$ with $L$.}$$
        we have $\Pi(\chi)\vDash r\_hh(h(l_1,t_\psi),h(\psi,t_\psi))$, $\dots$, and $\Pi(\chi)\vDash r\_hh(h(l_m,t_\psi),h(\psi,t_\psi))$. As ($c_1$), ($c_2$), and ($c_3$)---according to Proposition \ref{pro:direct_NESS_literal} and \ref{pro:complete_literal}---$(e,t)$ a direct NESS-cause of $(l_j,t_\psi)$ and thus $\Pi(\chi)\vDash direct\_ness(o(e,t),h(l_j,t_\psi))$. Therefore, given the rule:
            \begin{align*}
                \eqref{eq:direct_formula}: direct\_ness(Event,h(GD,T))~:-~&direct\_ness(Event,h(L,T)),\\ &r\_hh(h(L,T),h(GD,T)).
            \end{align*}
        we have $\Pi(\chi)\vDash direct\_ness(o(e,t),h(\psi,t_\psi))$.
    \end{proof}

    \begin{proposition}[Completeness if $\psi$ a DNF]\label{pro:complete_disjunction}
        Given $(e,t)\in\mathbb{E}\times\mathbb{T}$ a direct NESS-cause of $(\psi,t_\psi)$, where $\psi = \psi_1 \vee \dots \vee \psi_m$ a DNF, subsumption minimal, and tautology free formula of $\mathcal{P}$ such that $\psi\not\vDash\bot$, $\Pi(\chi)\vDash direct\_ness(o(e,t),h(\psi,t_\psi))$.
    \end{proposition}

    \begin{proof}
        Given $(e,t)$ a direct NESS-cause of $(\psi,t_\psi)$---according to Proposition \ref{pro:direct_NESS_DNF}---($c_1$) $\exists j\in \left\{1,\dots,m\right\},~S^{\chi}(t_\psi)\vDash \psi_j$, ($c_2$) $S^{\chi}(t)\triangleright \left\{e\right\}\vDash l\in\psi_j$, ($c_3$) $\forall t',~t<t'\leq t_\psi$, $S^{\chi}(t')\vDash l$, and ($c_4$) $S^{\chi}(t)\not\vDash l$.
        
        Considering the disjunctive normal form $\psi = \psi_1 \vee \dots \vee \psi_m$ true at $t_\psi$, $\Pi(\chi)\vDash disj(\psi)$, $\Pi(\chi)\vDash in(\psi,\psi_1)$, $\dots$, $\Pi(\chi)\vDash in(\psi,\psi_m)$, and $\Pi(\chi)\vDash holds(\psi,t_\psi)$. Additionally, considering ($c_1$), $\Pi(\chi)\vDash holds(\psi_j,t_\psi)$. Thus, given the rule:
        $$\eqref{eq:r_hh_disj}: r\_hh(h(GD\_L,T),h(D,T))~:-~disj(D), holds(D,T), in(D,GD\_L), holds(GD\_L,T).$$
        we have $\Pi(\chi)\vDash r\_hh(h(\psi_j,t_\psi),h(\psi,t_\psi))$. As ($c_2$), ($c_3$), and ($c_4$)---according to Proposition \ref{pro:direct_NESS_conjunction} and \ref{pro:complete_conjunction}---$(e,t)$ a direct NESS-cause of $(\psi_j,t_\psi)$ and thus $\Pi(\chi)\vDash direct\_ness(o(e,t),h(\psi_j,t_\psi))$. Therefore, given the rule:
            \begin{align*}
                \eqref{eq:direct_formula}: direct\_ness(Event,h(GD,T))~:-~&direct\_ness(Event,h(L,T)),\\ &r\_hh(h(L,T),h(GD,T)).
            \end{align*}
        we have $\Pi(\chi)\vDash direct\_ness(o(e,t),h(\psi,t_\psi))$.
    \end{proof}

    \begin{proposition}[Soundness if $\psi$ a literal]\label{pro:sound_literal}
        Given $(e,t)\in\mathbb{E}\times\mathbb{T}$ and $\Pi(\chi)\vDash direct\_ness(o(e,t),$ $h(l,t_l))$, $(e,t)$ is a direct NESS-cause of $(l,t_l)$.
    \end{proposition}
    
    \begin{proof}
        We consider $(e,t)$ such that $\Pi(\chi)\vDash\rho$ where $\rho=direct\_ness(o(e,t),h(l,t_l))$. Then---according to Proposition \ref{pro:ASP}---given that $\exists S\in AS(\Pi(\chi))$ such that $\rho\in S$, $\exists r\in\Pi(\chi)$, $head(r)=\rho$, $body^{+}(r)\subseteq S$, and $body^{-}(r)\cap S=\varnothing$.
        
        We denote $t_l=t+n$ with $n\in\mathbb{N}^*$ given that we model time in a discretised way and that $n\leq0$ would contradict $t_l> t$. Let $P(n)$ be the proposition: given $(e,t)\in\mathbb{E}\times\mathbb{T}$ and $\Pi(\chi)\vDash direct\_ness(o(e,t),h(l,t+n))$, $(e,t)$ is a direct NESS-cause of $(l,t+n)$. We give a proof by induction on $n$.
        
        \textit{Base case:} In the situation where $n=1$, two cases are possible: either $t=-1$, or $t\geq 0$. We will consider both cases.
        
        Case 1: when $t=-1$, $(l,t+n)$ is $(l,0)$, thus $\rho=direct\_ness(o(e,-1),h(l,0))$. The only rule $r$ that may have been used to derive $\rho$ is:
        $$\eqref{eq:direct_S0}: direct\_ness(o(ini(l),-1),h(l,0))~:-~initially(l).$$
        Hence, we necessarily have $e=ini_l$ and given $body^{+}(r)\subseteq S$, we have $initially(l)\in S$, thus $l\in S_0$. Therefore, by our action language semantics $ini_l\in E^{\chi}(-1)$ and $S^{\chi}(-1)\triangleright \left\{ini_l\right\} \vDash l$. Additionally, as $t=-1$ and $t+n=0$, $\forall t', ~-1<t'\leq 0, ~S^{\chi}(t')\vDash l$ is satisfied given that $S^{\chi}(0)\vDash l$. Finally, $S^{\chi}(-1)\not\vDash l$ is satisfied by the fact that $S^{\chi}(-1)=\varnothing$. Therefore---according to Proposition \ref{pro:direct_NESS_literal}---$(ini_l,-1)$ is a direct NESS-cause of $(l,0)$.
        
        Case 2: $(l,t+n)$ is $(f,t+1)$, or $(\neg f,t+1)$, thus $\rho=direct\_ness(o(e,t),h(f,t+1))$, or $\rho=direct\_ness(o(e,t),h(neg(f),t+1))$. For each case a unique rule $r$ may have been used to derive $\rho$. Those rules are:
        $$\eqref{eq:direct_ini}: direct\_ness(o(e,t),h(f,t+1))~:-~initiated(e,f,t).$$
        $$\eqref{eq:direct_ter}: direct\_ness(o(e,t),h(neg(f),t+1))~:-~terminated(e,f,t).$$
        Given $body^{+}(r)\subseteq S$, we have either $initiated(e,f,t)\in S$ or $terminated(e,f,t)\in S$, thus $S^{\chi}(t)\triangleright \left\{e\right\}\vDash l$ and $S^{\chi}(t)\not\vDash l$. Additionally, given $t_l=t+1$, $\forall t', ~t<t'\leq t+1, ~S^{\chi}(t')\vDash l$ is satisfied given that $S^{\chi}(t)\triangleright \left\{e\right\}\vDash l$ and $e\in E^{\chi}(t)$. Therefore---according to Proposition \ref{pro:direct_NESS_literal}---$(e,t)$ is a direct NESS-cause of $(l,t+1)$. $P(1)$ is thus true for both cases.
        
        \textit{Induction step:} We now show that for every $k\geq1$, if our induction hypothesis $P(k)$ holds---given $(e,t)\in\mathbb{E}\times\mathbb{T}$ and $\Pi(\chi)\vDash direct\_ness(o(e,t),h(l,t+k))$, $(e,t)$ is a direct NESS-cause of $(l,t+k)$---then $P(k+1)$ also holds---given $(e,t)\in\mathbb{E}\times\mathbb{T}$ and $\Pi(\chi)\vDash direct\_ness(o(e,t),h(l,t+k+1))$, $(e,t)$ is a direct NESS-cause of $(l,t+k+1)$.
        
        Two rules $r$ can be used to derive $\rho=direct\_ness(o(e,t),h(l,t+k+1))$ given $k=n\in\mathbb{N}^*$. 
        
        Case 1:
            \begin{align*}
                \eqref{eq:direct_formula}: direct\_ness(o(e,t),h(l,t+k+1))~:-~&direct\_ness(o(e,t),h(l',t+k+1)),\\ &r\_hh(h(l',t+k+1),h(l,t+k+1)).
            \end{align*}
        In this case, given $body^{+}(r)\subseteq S$, we should have $r\_hh(h(l',t+k+1),h(l,t+k+1))$. This is impossible given that $r\_hh(h(l',t+k+1),h(l,t+k+1))$ can only be derived by rules \eqref{eq:r_hh_conj} and \eqref{eq:r_hh_disj} which require either $l$ to be a conjunction---$conj(l)$---or a disjunction---$disj(l)$.
        
        Case 2:
            \begin{align*}
                \eqref{eq:direct_inertia}: direct\_ness(o(e,t),h(l,t+k+1))~:-~&direct\_ness(o(e,t),h(l,t+k)),\\ &inertia(h(l,t+k),h(l,t+k+1)).
            \end{align*}
        In this case, given $body^{+}(r)\subseteq S$, we have $direct\_ness(o(e,t),h(l,t+k))\in S$ and $inertia(h(l,t+k),h(l,t+k+1))\in S$. Thus, according to our induction hypothesis $(e,t)$ is a direct NESS-cause of $(l,t+k)$. According to Proposition \ref{pro:direct_NESS_literal}, this means $S^{\chi}(t)\triangleright \left\{e\right\}\vDash l$, $\forall t',~t<t'\leq t+k$, $S^{\chi}(t')\vDash l$, and $S^{\chi}(t)\not\vDash l$. Additionally, as $inertia(h(l,t+k),h(l,t+k+1))\in S$ and the rule:
        $$\eqref{eq:inertia}: inertia(h(l,t+k),h(l,t+k+1))~:-~holds(l,t+k), holds(l,t+k+1), literal(l).$$
        we can deduce that $S^{\chi}(t+k+1)\vDash l$, thus $\forall t',~t<t'\leq t+k+1$, $S^{\chi}(t')\vDash l$. It follows from Proposition \ref{pro:direct_NESS_literal} that $(e,t)$ is a direct NESS-cause of $(l,t+k+1)$.
        
        \textit{Conclusion:} Since the base case and the induction step have been proved as true, by mathematical induction the statement $P(n)$ holds for every $n\in\mathbb{N}^*$.
    \end{proof}

    \begin{proposition}[Soundness if $\psi$ a literal conjunction]\label{pro:sound_conjunction}
        Given $(e,t)\in\mathbb{E}\times\mathbb{T}$ and $\Pi(\chi)\vDash direct\_ness(o(e,t),h(\psi,t_\psi))$, where $\psi = l_1 \wedge \dots \wedge l_m$, $(e,t)$ is a direct NESS-cause of $(\psi,t_\psi)$.
    \end{proposition}

    \begin{proof}
        The only rule $r$ that may have been used to derive $\rho=direct\_ness(o(e,t),h(\psi,t_\psi))$ given $\psi = l_1 \wedge \dots \wedge l_m$ is:
        $$\eqref{eq:direct_formula}: direct\_ness(o(e,t),h(\psi,t_\psi))~:-~direct\_ness(o(e,t),h(l_j,t_\psi)), r\_hh(h(l_j,t_\psi),h(\psi,t_\psi)).$$
        In this case, given $body^{+}(r)\subseteq S$, we have $direct\_ness(o(e,t),h(l_j,t_\psi))\in S$ and $r\_hh(h(l_j,t_\psi),$ $h(\psi,t_\psi))\in S$. Therefore---according to Proposition \ref{pro:sound_literal} and \ref{pro:direct_NESS_literal}---$(e,t)$ is a direct NESS-cause of $(l_j,t_\psi)$ and thus $S^{\chi}(t)\triangleright \left\{e\right\}\vDash l_j$, $\forall t',~t<t'\leq t_\psi$, $S^{\chi}(t')\vDash l_j$, and $S^{\chi}(t)\not\vDash l_j$. Given $r\_hh(h(l_j,t_\psi),h(\psi,t_\psi))\in S$ and the rule:
        $$\eqref{eq:r_hh_conj}: r\_hh(h(l_j,t_\psi),h(\psi,t_\psi))~:-~conj(\psi), holds(\psi,t_\psi), in(\psi,l_j).$$
        we have $conj(\psi)\in S$, $holds(\psi,t_\psi)\in S$, and $in(\psi,l_j)\in S$. Therefore, $j\in \left\{1,\dots,m\right\}$ and thus---according to Proposition \ref{pro:direct_NESS_conjunction}---$(e,t)$ is a direct NESS-cause of $(\psi,t_\psi)$.
    \end{proof}

    \begin{proposition}[Soundness if $\psi$ a DNF]\label{pro:sound_disjunction}
        Given $(e,t)\in\mathbb{E}\times\mathbb{T}$ and $\Pi(\chi)\vDash direct\_ness(o(e,t),$ $h(\psi,t_\psi))$, where $\psi = \psi_1 \vee \dots \vee \psi_m$ a DNF, subsumption minimal, and tautology free formula of $\mathcal{P}$ such that $\psi\not\vDash\bot$, $(e,t)$ is a direct NESS-cause of $(\psi,t_\psi)$.
    \end{proposition}

    \begin{proof}
        As above, the only rule $r$ that may have been used to derive $\rho=direct\_ness(o(e,t),$ $h(\psi,t_\psi))$ given $\psi = \psi_1 \vee \dots \vee \psi_m$ is:
        $$\eqref{eq:direct_formula}: direct\_ness(o(e,t),h(\psi,t_\psi))~:-~direct\_ness(o(e,t),h(\psi_j,t_\psi)), r\_hh(h(\psi_j,t_\psi),h(\psi,t_\psi)).$$
        In this case, given $body^{+}(r)\subseteq S$, we have $direct\_ness(o(e,t),h(\psi_j,t_\psi))\in S$ and $r\_hh(h(\psi_j,t_\psi),$ $h(\psi,t_\psi))\in S$. Therefore---according to Proposition \ref{pro:sound_conjunction} and \ref{pro:direct_NESS_conjunction}---$(e,t)$ is a direct NESS-cause of $(\psi_j,t_\psi)$ and thus given $\psi_j=l_1 \wedge \dots \wedge l_n$, $\exists i\in \left\{1,\dots,n\right\},~S^{\chi}(t)\triangleright \left\{e\right\}\vDash l_i$, $\forall t',~t<t'\leq t_\psi$, $S^{\chi}(t')\vDash l_i$, and $S^{\chi}(t)\not\vDash l_i$. Given $r\_hh(h(\psi_j,t_\psi),h(\psi,t_\psi))\in S$ and the rule:
        $$\eqref{eq:r_hh_disj}: r\_hh(h(\psi_j,t_\psi),h(\psi,t_\psi))~:-~disj(\psi), holds(\psi,t_\psi), in(\psi,\psi_j), holds(\psi_j,t_\psi).$$
        we have $disj(\psi)\in S$, $holds(\psi,t_\psi)\in S$, $in(\psi,\psi_j)\in S$, and $holds(\psi_j,t_\psi)\in S$. Therefore, $j\in \left\{1,\dots,m\right\}$ and $S^{\chi}(t_\psi)\vDash\psi_j$, and thus---according to Proposition \ref{pro:direct_NESS_DNF}---$(e,t)$ is a direct NESS-cause of $(\psi,t_\psi)$.
    \end{proof}

    \begin{theorem}[direct NESS completeness and soundness]
        Given a causal setting $\chi$ and a DNF, subsumption minimal, and tautology free formula $\psi\in\mathcal{P}$ such that $\psi\not\vDash\bot$, $(e,t)\in\mathbb{E}\times\mathbb{T}$ is a direct NESS-cause of $(\psi,t_\psi)$, iff $\Pi(\chi)\vDash direct\_ness(o(e,t),h(\psi,t_\psi))$.
    \end{theorem}
    
    \begin{proof}
        Theorem \ref{the:complete_and_sound} follows from Propositions \ref{pro:complete_literal} to \ref{pro:sound_disjunction}. 
    \end{proof}

\section*{Proofs of Completeness and Soundness of NESS-Causes}
    
    \begin{theorem}[NESS completeness and soundness]
        Given a causal setting $\chi$ and a DNF, subsumption minimal, and tautology free formula $\psi\in\mathcal{P}$ such that $\psi\not\vDash\bot$, $(e,t)\in\mathbb{E}\times\mathbb{T}$ is a NESS-cause of $(\psi,t_\psi)$, iff $\Pi(\chi)\vDash ness(o(e,t),h(\psi,t_\psi))$.
    \end{theorem}
    
    \begin{proof}
        $[\implies]$ 
            Let $(e,t)$ be a NESS-cause of $(\psi,t_{\psi})$. By Definition \ref{def:NESS}, $\exists C'\subseteq\mathbb{E}\times\mathbb{T}$ such that $(e,t)\in C'$ where $C'$ a sufficient set of NESS-causes of $(\psi,t_\psi)$ and $\exists C\subseteq\mathbb{E}\times\mathbb{T}$ such that $C\rightsquigarrow(\psi,t_{\psi})$. Two possibilities are available to us: either we are in the base case, or in the recursive case of Definition \ref{def:NESS}.
            
            Base Case: In this particular case where $C'=C$, $(e,t)\in C$. Thus---according to Definition \ref{def:direct_NESS}---$(e,t)$ is a direct NESS-cause of $(\psi,t_{\psi})$. Therefore---according to Theorem \ref{the:complete_and_sound}---$\Pi(\chi)\vDash direct\_ness(o(e,t),h(\psi,t_\psi))$. Then, given rule:
            $$\eqref{eq:direct_is_ness}: ness(o(E1,T1),h(GD\_L,T2)) :-~direct\_ness(o(E1,T1),h(GD\_L,T2)).$$
            we have $\Pi(\chi)\vDash ness(o(e,t),h(\psi,t_\psi))$.
            
            Recursive Case: In this case where $C'\neq C$ we know by Definition \ref{def:NESS} that there is a set $C_R=C\setminus C'$ of `removable' occurrence of events whose partition matches the decreasing sequence $t_1,\dots,t_k$---$C_R(t_1),\dots,C_R(t_k)$---and a set of `overwhelming' occurrence of events $C_O=C'\setminus C$ decomposable in a sequence of subsets $C_{O_{1}},\dots,C_{O_{k}}$ such that:
                \begin{itemize}
                    \item[-] $C_O=\bigcup_{i\in \left\{1,\dots,k\right\}}C_{O_{i}}$.
                    \item[-] $\forall i\in \left\{1,\dots,k\right\}, C_R(t_i)=\varnothing\implies C_{O_{i}}=\varnothing$.
                    \item[-] $\forall i\in \left\{1,\dots,k\right\}, C_R(t_i)\neq\varnothing\implies C_{O_{i}}$ is a sufficient set of NESS-causes of $(tri(C_R(t_i)),t_{i})$.
                \end{itemize}
            Given this decomposition, we can write $C'=C_O\cup(C\cap C')$ and $C=C_R\cup(C\cap C')$. The case where $(e,t)\in (C\cap C')$ having the same proof as the base case above, we will now consider the cases where $(e,t)\in C_O$. More precisely and given $C_O=\bigcup_{i\in \left\{1,\dots,k\right\}}C_{O_{i}}$, we are in the cases where $C_R(t_i)\neq\varnothing$ and $(e,t)\in C_{O_{i}}$, thus $(e,t)$ is a NESS-causes of $(tri(C_R(t_i)),t_{i})$. Therefore, $\exists (e',t_i)\in C_R(t_i)$, $(e,t)$ is a NESS-cause of $(tri(e'),t_i)$. Additionally, as $(e',t_i)\in C_R(t_i)$ and $C=C_R\cup(C\cap C')$, $(e',t_i)\in C$ which means by the base case above that $\Pi(\chi)\vDash ness(o(e',t_i),h(\psi,t_\psi))$. Two cases can then be considered.
            
            In the first case corresponding to the base case, $(e,t)$ is a direct NESS-cause of $(tri(e'),t_i)$, thus---according to Theorem \ref{the:complete_and_sound}---$\Pi(\chi)\vDash ness(o(e,t),h(tri(e'),t_i))$. Then, given rules:
            \begin{align*}
                \eqref{eq:direct_is_ness}: ness(o(E1,T1),h(GD\_L,T3)) :-~ &actual(o(E1,T1),o(E2,T2)),\\
                &ness(o(E2,T2),h(GD\_L,T3)).\\
                \eqref{eq:actual_ness}: actual(o(E1,T1),o(E2,T2)) :-~ &ness(o(E1,T1),h(GD,T2)),\\
                &happens(E2,GD,T2), auto(E2,GD,Eff).
            \end{align*}
            we have $\Pi(\chi)\vDash actual(o(e,t),o(e',t_i))$, and then $\Pi(\chi)\vDash ness(o(e,t),h(\psi,t_\psi))$.
            
            The second case corresponds to the recursive case where $(e,t)$ is a NESS-cause of $(tri(e''),t_j)$ with $(e'',t_j)$ a direct NESS-cause of $(tri(e'),t_i)$. Given that we have a bounded past formalisation, applying the same reasoning recursively will inevitably lead us to the above first case. We will thus be able to say that $\Pi(\chi)\vDash actual(o(e,t),o(e'',t_j))$. Then, we can chain forward by applying rules \eqref{eq:direct_is_ness} and \eqref{eq:actual_ness} until deriving $\Pi(\chi)\vDash ness(o(e,t),h(\psi,t_\psi))$.
            
        $[\impliedby]$
            We consider $(e,t)$ such that $\Pi(\chi)\vDash\rho$ where $\rho=ness(o(e,t),h(\psi,t_{\psi}))$. Then---according to Proposition \ref{pro:ASP}---given that $\exists S\in AS(\Pi(\chi))$ such that $\rho\in S$, $\exists r\in\Pi(\chi)$, $head(r)=\rho$, $body^{+}(r)\subseteq S$, and $body^{-}(r)\cap S=\varnothing$.
        
            We denote $t_{\psi}=t+n$ with $n\in\mathbb{N}^*$. Let $P(n)$ be the proposition: given a causal setting $\chi$, a DNF, subsumption minimal, and tautology free formula $\psi\in\mathcal{P}$ such that $\psi\not\vDash\bot$ and $\Pi(\chi)\vDash ness(o(e,t),h(\psi,t+n))$, $(e,t)\in\mathbb{E}\times\mathbb{T}$ is a NESS-cause of $(\psi,t+n)$. We give a proof by induction on $n$.
            
            \textit{Base case:} In the situation where $n=1$, $(\psi,t+n)$ is $(\psi,t+1)$, thus $\rho=ness(o(e,t),h(\psi,t+1))$. For this case the only rule $r$ that may have been used to derive $\rho$ is:
            $$\eqref{eq:direct_is_ness}: ness(o(e,t),h(\psi,t+1))~:-~direct\_ness(o(e,t),h(\psi,t+1)).$$
            Case of rule \eqref{eq:direct_is_ness}: given $body^{+}(r)\subseteq S$, we have $direct\_ness(o(e,t),h(\psi,t+1))\in S$, thus---according to Theorem \ref{the:complete_and_sound}---$\exists C\subseteq\mathbb{E}\times\mathbb{T}$ such that $(e,t)\in C$ and $C\rightsquigarrow(\psi,t+1)$. According to Definition \ref{def:NESS} base case, by taking $C'=C$, the occurrence of events set $C'$ is a sufficient set of NESS-causes of $(\psi,t+1)$, Additionally, given $(e,t)\in C$ and $C'=C$, $(e,t)\in C'$. Hence, $(e,t)$ is a NESS-cause of $(\psi,t+1)$. $P(1)$ is thus true.
           
            \textit{Induction step:} We now show that for every $k\geq1$, if our induction hypothesis $P(k)$ holds---given a causal setting $\chi$, a DNF, subsumption minimal, and tautology free formula $\psi\in\mathcal{P}$ such that $\psi\not\vDash\bot$, and $\Pi(\chi)\vDash ness(o(e,t),h(\psi,t+k))$, $(e,t)\in\mathbb{E}\times\mathbb{T}$ is a NESS-cause of $(\psi,t+k)$---then $P(k+1)$ also holds---given a causal setting $\chi$, a DNF, subsumption minimal, and tautology free formula $\psi\in\mathcal{P}$ such that $\psi\not\vDash\bot$, and $\Pi(\chi)\vDash ness(o(e,t),h(\psi,t+k+1))$, $(e,t)\in\mathbb{E}\times\mathbb{T}$ is a NESS-cause of $(\psi,t+k+1)$. Rules \eqref{eq:direct_is_ness} and \eqref{eq:ness_trig} can be used to derive $\rho=ness(o(e,t),h(\psi,t+k+1))$ given $k=n\in\mathbb{N}^*$. 
            
            Case of rule \eqref{eq:direct_is_ness}:
            $$\eqref{eq:direct_is_ness}: ness(o(e,t),h(\psi,t+k+1))~:-~direct\_ness(o(e,t),h(\psi,t+k+1)).$$
            given $body^{+}(r)\subseteq S$, we have $direct\_ness(o(e,t),h(\psi,t+k+1))\in S$, thus---according to Theorem \ref{the:complete_and_sound}---$\exists C\subseteq\mathbb{E}\times\mathbb{T}$ such that $(e,t)\in C$ and $C\rightsquigarrow(\psi,t+k+1)$. According to Definition \ref{def:NESS} base case, by taking $C'=C$, the occurrence of events set $C'$ is a sufficient set of NESS-causes of $(\psi,t+k+1)$, Additionally, given $(e,t)\in C$ and $C'=C$, $(e,t)\in C'$. Hence, $(e,t)$ is a NESS-cause of $(\psi,t+k+1)$.
            
            Case of rule \eqref{eq:ness_trig}:
            $$\eqref{eq:ness_trig}: ness(o(e,t),h(\psi,t+k+1))~:-~actual(o(e,t),o(e',t')), ness(o(e',t'),h(\psi,t+k+1)).$$
            given $body^{+}(r)\subseteq S$, we have $ness(o(e',t'),h(\psi,t+k+1))\in S$ and $actual(o(e,t),o(e',t'))\in S$. Because of our assumption that causes precedes effects in time, we know that $t<t'<t+k+1$. The difference between $t'$ and $t+k+1$ is at most $t+k$. Therefore---according to our induction hypothesis $P(k)$---$(e',t')$ is a NESS-cause of $(\psi,t+k+1))$. Additionally, the only rule $r$ that may have been used to derive $actual(o(e,t),o(e',t'))$ is the rule:
            $$\eqref{eq:actual_ness}: actual(o(e,t),o(e',t'))~:-~ness(o(e,t),h(\psi',t')), happens(e',\psi',t'), auto(e',\psi',\phi').$$
            Given $body^{+}(r)\subseteq S$, we have $ness(o(e,t),h(\psi',t'))\in S$, $happens(e',\psi',t')\in S$, and $auto(e',\psi',\phi')\in S$. As above, we can deduce that $(e,t)$ is a NESS-cause of $(\psi',t')$, where $e'\in\mathbb{U}$, $tri(e')=\psi'$, and $e'\in E^{\chi}(t')$. Consequently---according to the recursive case of Definition \ref{def:NESS}---$(e,t)$ is a NESS-cause of $(\psi,t+k+1)$.

            \textit{Conclusion:} Since the base case and the induction step have been proved as true, by mathematical induction the statement $P(n)$ holds for every $n\in\mathbb{N}^*$.
    \end{proof}

\section*{Proofs of Completeness and Soundness of Actual Causes}

    \begin{theorem}[actual cause completeness and soundness]
        Given a causal setting $\chi$, a DNF, subsumption minimal, and tautology free formula $\psi\in\mathcal{P}$ such that $\psi\not\vDash\bot$, and $tri(e')=\psi$, $(e,t)\in\mathbb{E}\times\mathbb{T}$ is an actual cause of $(e',t')\in\mathbb{E}\times\mathbb{T}$, iff $\Pi(\chi)\vDash actual(o(e,t),o(e',t'))$.
    \end{theorem}
    
    \begin{proof}
        $[\implies]$ 
            Let $(e,t)$ be an actual cause of $(e',t')$. By Definition \ref{def:actual}, $(e,t)$ is a NESS-cause of $(tri(e'),t')$, thus---according to Theorem \ref{the:complete_and_sound_1}---$\Pi(\chi)\vDash ness(o(e,t),h(tri(e'),t'))$. As $e'\in E^{\chi}(t')$ and $e\in\mathbb{U}$, $\Pi(\chi)\vDash happens(e',tri(e'),t')$ and $\Pi(\chi)\vDash auto(e',tri(e'),eff(e'))$. Hence, given the rule:
                \begin{align*}
                    \eqref{eq:actual_ness}: actual(o(E1,T1),o(E2,T2))~:-~&ness(o(E1,T1),h(GD,T2)), happens(E2,GD,T2),\\ &auto(E2,GD,Eff).
                \end{align*}
            we have $\Pi(\chi)\vDash actual(o(e,t),o(e',t'))$.
            
        $[\impliedby]$
            We consider $(e,t)$ and $(e',t')$ such that $\Pi(\chi)\vDash\rho$ where $\rho=actual(o(e,t),o(e',t'))$. Then---according to Proposition \ref{pro:ASP}---given that $\exists S\in AS(\Pi(\chi))$ such that $\rho\in S$, $\exists r\in\Pi(\chi)$, $head(r)=\rho$, $body^{+}(r)\subseteq S$, and $body^{-}(r)\cap S=\varnothing$.
            
            The only rule $r$ that may have been used to derive $\rho$ is:
                \begin{align*}
                    \eqref{eq:actual_ness}: actual(o(e,t),o(e',t'))~:-~&ness(o(e,t),h(tri(e'),t')), happens(e',tri(e'),t'),\\ &auto(e',tri(e'),eff(e')).
                \end{align*}
            Given $body^{+}(r)\subseteq S$, we have $ness(o(e,t),h(tri(e'),t'))\in S$, thus---according to Theorem \ref{the:complete_and_sound_1}---$(e,t)$ is a NESS-cause of $(tri(e'),t')$. Hence, by Definition \ref{def:actual}, $(e,t)$ is an actual cause of $(e',t')$.
    \end{proof}

\section*{Entire ASP Code for Examples \ref{ex:preemption} and \ref{ex:duplication}}
    
    \subsection*{Program $\pi_{con}(\kappa)$}
    
    \begin{multicols}{2}
        \begin{flalign*}
            time(0..3).&&\\
            initially(t\_os).&&\\
            fluent(m\_k).&&\\
            fluent(w\_m).&&\\
            fluent(e\_n).&&\\
            fluent(w\_s).&&\\
            fluent(t\_os).&&\\
            fluent(s\_sup).&&\\
            fluent(d).&&
        \end{flalign*}
        
        \begin{flalign*}
            action(prod\_m,true,prod\_mEff).&&\\
            conj(prod\_mEff).&&\\
            in(prod\_mEff,m\_k).&&\\
            in(prod\_mEff,w\_m).&&
        \end{flalign*}
        
        \begin{flalign*}
            action(prod\_s,true,prod\_sEff).&&\\
            conj(prod\_sEff).&&\\
            in(prod\_sEff,e\_n).&&\\
            in(prod\_sEff,w\_s).&&\\
        \end{flalign*}
        
        \begin{flalign*}
            auto(dis\_w,dis\_wCond,dis\_wEff).&&\\
            disj(dis\_wCond).&&\\
            in(dis\_wCond,w\_s).&&\\
            in(dis\_wCond,dis\_wCond1).&&\\
            conj(dis\_wCond1).&&\\
            in(dis\_wCond1,w\_m).&&\\
            in(dis\_wCond1,t\_os).&&\\
            conj(dis\_wEff).&&\\
            in(dis\_wEff,s\_sup).&&\\
        \end{flalign*}
        
        \begin{flalign*}
            auto(fau\_p,fau\_pCond,fau\_pEff).&&\\
            conj(fau\_pCond).&&\\
            in(fau\_pCond,s\_sup).&&\\
            conj(fau\_pEff).&&\\
            in(fau\_pEff,d).&&
        \end{flalign*}
    \end{multicols}
    
    \subsection*{Program $\pi_{sce}(\sigma)$}
    \begin{multicols}{2}
        \subsubsection*{Example \ref{ex:preemption}}
        \begin{align*}
            performs(prod\_m,0).\\
            performs(prod\_s,1).
        \end{align*}
        
        \subsubsection*{Example \ref{ex:duplication}}
        \begin{align*}
            performs(prod\_m,0).\\
            performs(prod\_s,0).
        \end{align*}
    \end{multicols}
    
    \subsection*{Program $\pi_{\mathbb{A}}$}
        \subsubsection*{Events precondition axioms}
            \begin{align*}
                triggered(A,GD,T) :-~ &action(A,GD,Effect), performs(A,T), holds(GD,T).\\
                triggered(U,GD,T) :-~ &auto(U,GD,Effect), holds(GD,T).	\\
                overtaken(E1,T) :-~ &triggered(E1,\_,T), happens(E2,\_,T), priority(E2,E1), E2!=E1.\\
                happens(E,GD,T) :-~ &triggered(E,GD,T), not~overtaken(E,T).
            \end{align*}
        
        \subsubsection*{Events effect axioms}
            \begin{align*}
                apply(A,Effect,T) :-~ &happens(A,GD,T), action(A,GD,Effect).\\
                apply(U,Effect,T) :-~ &happens(U,GD,T), auto(U,GD,Effect).\\
                initiated(E,F,T) :-~ &apply(E,Effect,T), in(Effect,F), not~holds(F,T), fluent(F).\\
                terminated(E,F,T) :-~ &apply(E,Effect,T), in(Effect,neg(F)), holds(F,T), fluent(F).\\
                holds(F,0) :-~ &initially(F), fluent(F).\\
                holds(F,T+1) :-~ &initiated(E,F,T).\\
                holds(F,T+1) :-~ &holds(F,T), fluent(F), time(T),not~terminated(E,F,T):event(E).\\
                initially(neg(F)) :-~ &not~initially(F), fluent(F).\\
                holds(neg(F),T) :-~ &not~holds(F,T), fluent(F), time(T).
            \end{align*}
        
        \subsubsection*{Conjunction, disjunction and negation axioms}
            \begin{align*}
                negative(neg(F)) :-~ &fluent(F).\\
                literal(F) :-~ &fluent(F).\\
                literal(F) :-~ &negative(F).\\			
                in(L,L) :-~ &literal(L).\\
                holds(true,T) :-~ &time(T).\\
                holds(C,T) :-~ &conj(C), time(T), not~action(\_,\_,C), not~auto(\_,\_,C), holds(F,T):in(C,F).\\
                holds(D,T) :-~ &disj(D), holds(F,T), in(D,F), not~action(\_,\_,D), not~auto(\_,\_,D).\\
                event(A) :-~ &action(A,GD,Effect).\\
                event(U) :-~ &auto(U,GD,Effect).
            \end{align*}

    \subsection*{Program $\pi_{\mathbb{C}}$}
        \begin{align*}
            event(A,GD,Effect) :-~ &action(A,GD,Effect).\\
            event(U,GD,Effect) :-~ &auto(U,GD,Effect).\\
            inertia(h(L,T),h(L,T+1)) :-~ &holds(L,T), holds(L,T+1), literal(L).\\
            r\_hh(h(GD\_L,T),h(C,T)) :-~ &conj(C), holds(C,T), in(C,GD\_L).\\
            r\_hh(h(GD\_L,T),h(D,T)) :-~ &disj(D), holds(D,T), in(D,GD\_L), holds(GD\_L,T).
        \end{align*}   
        
        \subsubsection*{Direct NESS-Cause}
            \begin{align*}
                direct\_ness(o(ini(L),-1),h(L,0)) :-~ &initially(L).\\
                direct\_ness(o(E,T),h(F,T+1)) :-~ &initiated(E,F,T).\\
                direct\_ness(o(E,T),h(neg(F),T+1)) :-~ &terminated(E,F,T).\\
                direct\_ness(Event,h(L,T+1)) :-~ &direct\_ness(Event,h(L,T)),inertia(h(L,T),h(L,T+1)).\\
                direct\_ness(Event,h(GD,T)) :-~ &direct\_ness(Event,h(GD\_L,T)),r\_hh(h(GD\_L,T),h(GD,T)).
            \end{align*}
            
        \subsubsection*{NESS-Cause}
            \begin{align*}
                ness(o(E1,T1),h(GD\_L,T2)) :-~ &direct\_ness(o(E1,T1),h(GD\_L,T2)).\\
                ness(o(E1,T1),h(GD\_L,T3)) :-~ &actual(o(E1,T1),o(E2,T2)),ness(o(E2,T2),h(GD\_L,T3)).
            \end{align*}
            
        \subsubsection*{Actual Cause}
            \begin{align*}
                actual(o(E1,T1),o(E2,T2)) :-~ &ness(o(E1,T1),h(GD,T2)),happens(E2,GD,T2),auto(E2,GD,Eff).
            \end{align*}

\end{document}